\documentclass[pdflatex,sn-mathphys]{sn-jnl}% Math and Physical Sciences Reference Style
\usepackage{bbm}
\usepackage{subcaption}

\usepackage{lineno}
\newcommand*\patchAmsMathEnvironmentForLineno[1]{%
  \expandafter\let\csname old#1\expandafter\endcsname\csname #1\endcsname
  \expandafter\let\csname oldend#1\expandafter\endcsname\csname end#1\endcsname
  \renewenvironment{#1}%
     {\linenomath\csname old#1\endcsname}%
     {\csname oldend#1\endcsname\endlinenomath}}% 
\newcommand*\patchBothAmsMathEnvironmentsForLineno[1]{%
  \patchAmsMathEnvironmentForLineno{#1}%
  \patchAmsMathEnvironmentForLineno{#1*}}%
\AtBeginDocument{%
\patchBothAmsMathEnvironmentsForLineno{equation}%
\patchBothAmsMathEnvironmentsForLineno{align}%
\patchBothAmsMathEnvironmentsForLineno{flalign}%
\patchBothAmsMathEnvironmentsForLineno{alignat}%
\patchBothAmsMathEnvironmentsForLineno{gather}%
\patchBothAmsMathEnvironmentsForLineno{multline}%
}

\usepackage{latexsym}
\newcommand{\xmark}{\text{\sffamily X}}%

\jyear{2022}%

\theoremstyle{thmstyleone}%
\newtheorem{theorem}{Theorem}%  meant for continuous numbers
\theoremstyle{thmstyletwo}%

\theoremstyle{thmstylethree}%

\def\std{\sigma_{\mathrm{std}}}

\newcommand{\abs}[1]{\left\lvert#1\right\rvert}
\def\E{\mathrm{I\!E}}

\begin{document}

\title[Training Integer-Only Deep Recurrent  Neural Networks]{Training Integer-Only Deep Recurrent  Neural Networks
%\footnote[]{$\dag$ This article is an extended version of the paper selected as the winner of the best industry paper award of ICPRAM \cite{Sari_iRNN_ICPRAM2022}.}
}

%%=============================================================%%
%% Prefix	-> \pfx{Dr}
%% GivenName	-> \fnm{Joergen W.}
%% Particle	-> \spfx{van der} -> surname prefix
%% FamilyName	-> \sur{Ploeg}
%% Suffix	-> \sfx{IV}
%% NatureName	-> \tanm{Poet Laureate} -> Title after name
%% Degrees	-> \dgr{MSc, PhD}
%% \author*[1,2]{\pfx{Dr} \fnm{Joergen W.} \spfx{van der} \sur{Ploeg} \sfx{IV} \tanm{Poet Laureate} 
%%                 \dgr{MSc, PhD}}\email{iauthor@gmail.com}
%%=============================================================%%

\author*[1,2]{\fnm{Vahid} \sur{Partovi~Nia}}\email{vahid.partovinia@huawei.com}

\author[1]{\fnm{Eyy\"ub} \sur{Sari}}\email{eyyub.sari@gmail.com}
%\equalcont{These authors contributed equally to this work.}

\author[1]{\fnm{Vanessa} \sur{Courville}}\email{vanessa.courville@huawei.com}
%\equalcont{These authors contributed equally to this work.}

\author[3]{\fnm{Masoud} \sur{Asgharian}}\email{masoud.asgharian2@mcgill.ca}

\affil*[1]{\orgdiv{Huawei Noah's Ark Lab}, \orgname{Montreal Research Centre}, \orgaddress{\street{7101 Park Avenue}, \city{Montreal}, \postcode{H3N 1X9}, \state{Quebec}, \country{Canada}}}

\affil[2]{\orgdiv{Department of Mathematics and Industrial Engineering}, \orgname{Polytechnique Montreal}, \orgaddress{\street{2500 Chem. Polytechnique}, \city{Montreal}, \postcode{H3T 1J4}, \state{Quebec}, \country{Canada}}}

\affil[3]{\orgdiv{Department of Mathematics and Statistics}, \orgname{McGill University}, \orgaddress{\street{805 Sherbrooke Street West}, \city{Montreal}, \postcode{H3A 0B9}, \state{Quebec}, \country{Canada}}}

%%==================================%%
%% sample for unstructured abstract %%
%%==================================%%

\abstract{Recurrent neural networks (RNN) are the backbone of many text and speech applications. These architectures are typically made up of several computationally complex components such as; non-linear activation functions, normalization, bi-directional dependence and attention. In order to maintain good accuracy, these components are frequently run using full-precision floating-point computation, making them slow, inefficient and difficult to deploy on edge devices. In addition, the complex nature of these operations makes them challenging to quantize using standard quantization methods without a significant performance drop. We present a quantization-aware training method for obtaining a highly accurate integer-only recurrent neural network (iRNN). Our approach supports layer normalization, attention, and an adaptive piecewise linear (PWL) approximation of activation functions, to serve a wide range of state-of-the-art RNNs. The proposed method enables RNN-based language models to run on edge devices with $2\times$ improvement in runtime, and $4\times$ reduction in model size while maintaining similar accuracy as its full-precision counterpart.}


%%\pacs[JEL Classification]{D8, H51}

%%\pacs[MSC Classification]{35A01, 65L10, 65L12, 65L20, 65L70}

\maketitle
\section{Introduction}
Recurrent Neural Network (RNN) \cite{rumelhart1986rnn} architectures such as Long-Short Term Memory (LSTM) \cite{hochreiter1997lstm} or Gated Recurrent Units (GRU) \cite{cho2014gru} are the foundation of language modeling applications. They can be found in most large-scale systems such as neural machine translation   \cite{chen2018translstm,wang2019accelerating}, sentiment analysis \cite{DBLP:journals/corr/abs-1801-07883}, image captioning \cite{You_2016_CVPR,Mao-mRNN} and on-device systems such as Automatic Speech Recognition (ASR) \cite{he2019streaming}. Despite their pervasive use in academia and industry, RNN architectures require more elaborated studies in order to improve inference efficiency on-device. 

Recently, Transformer-based models \cite{vaswani2017transformer,devlin2019bert} have been proposed as an alternative to RNNs. Transformer architectures use an encoder-decoder model, where the encoder maps an input data sequence to learned vector representation, while the decoder uses the learned data to produce a single output. Transformers differ from RNNs because sequential data (such as words in a sentence) can be processed through the transformer's encoder concurrently, allowing for parallel training, not present in RNN models. However, RNNs, especially LSTMs, are still powerful tools for building highly accurate industry models. The combination of LSTM and transformer architectures has provided impressive results recently; \cite{chen2018translstm}, and \cite{wang2019accelerating} proposed an encoder-decoder neural machine translator which harvests the inherent parallelism of transformers for its encoder while utilizing LSTMs for a faster decoder. The LSTM decoder achieved a faster inference runtime than a transformer decoder in these studies. It was shown, for these cases, that Transformers may lose their parallelization capabilities when generating outputs based on conditional inputs. Moreover, they also suffer from quadratic computation complexity in terms of the sequence length \cite{wang2019accelerating}. \cite{kasai2021t2r} took transformers one step further and fine-tuned a pre-trained transformer into RNNs, benefiting the best of the two worlds. 

When running these models on edge devices, such as mobile or IoT devices, the physical hardware constraints must also be taken into consideration. In many of these devices, the number of computing cores is limited to a handful of processors, in which parallel-friendly transformer-based models lose their advantage. There have been several studies on quantizing transformers to adapt them for edge devices, but RNNs are largely ignored. Deploying RNN-based chatbot, conversational agent, and ASR on edge devices with limited memory and energy requires further computational improvements. The  8-bit integer neural networks quantization \cite{jacob2017quantization} for convolutional architectures (CNNs) is shown to be an almost free lunch approach to tackle the memory, energy, and latency costs, with a negligible accuracy drop \cite{krishnamoorthi2018quantizing}.

\begin{figure*}
\centering
\includegraphics[scale=0.65]{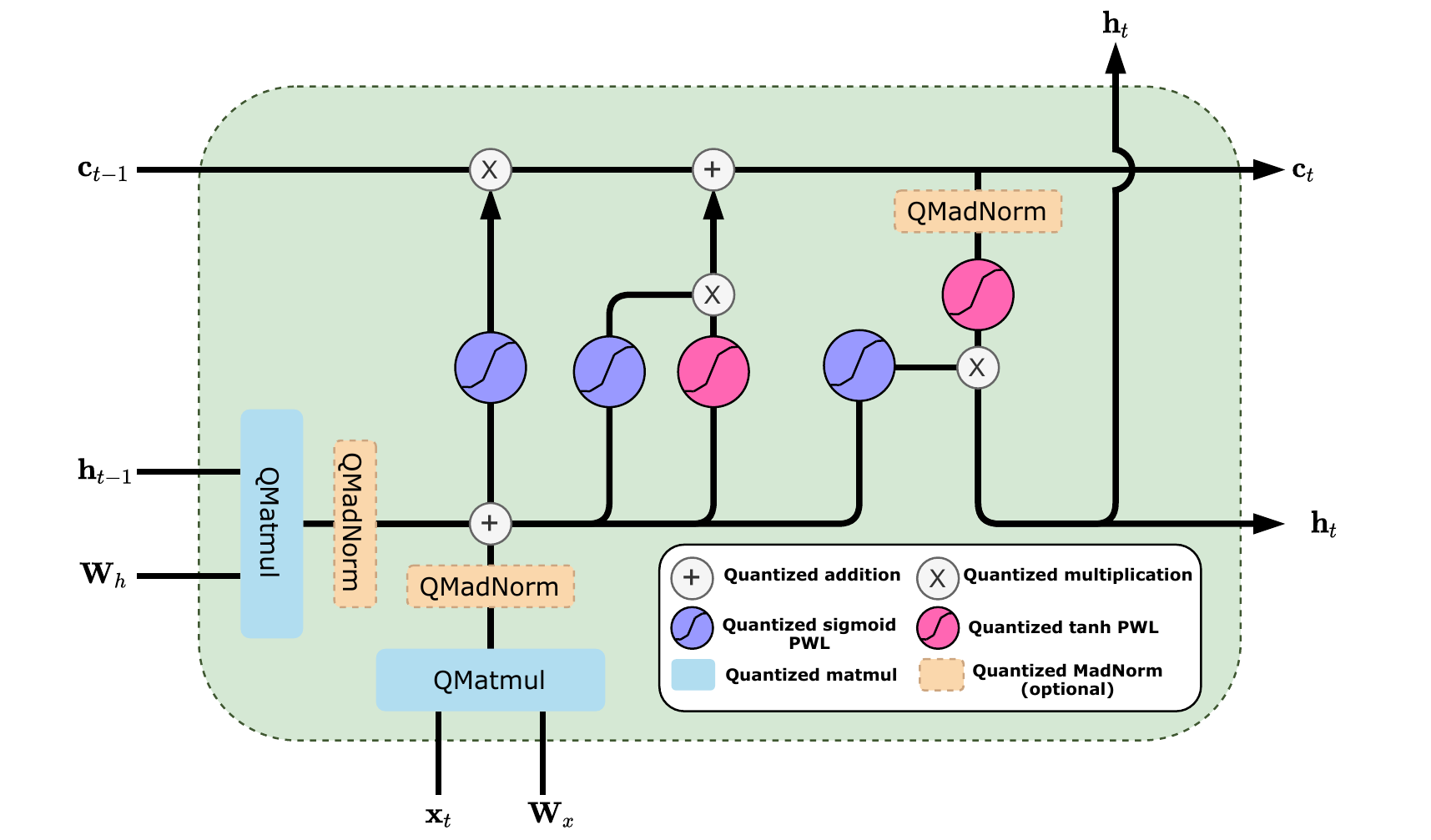}
\caption{Example of an integer-only LSTM cell (iLSTM). Layer normalization changes to quantized integer friendly MadNorm (QMadNorm), full-precision matrix multiplications change to integer matrix multiplications (QMatmul), sigmoid, and tanh activation functions are replaced by their corresponding piecewise linear (PWL) approximations. 
}
\label{fig:quantized-lstm}
\end{figure*}

Intuitively, quantizing RNNs is more challenging because the errors introduced by quantization will propagate in two directions, i) to the successive layers, like feedforward networks ii) across timesteps. Furthermore, RNN cells are computationally more complex;  they include several element-wise additions and multiplications. They also have different activation functions that rely on the exponential function, such as sigmoid and hyperbolic tangent (tanh).

Accurate fully-integer RNNs call for a new cell built using integer-friendly operations. Our primary motivation is to enable integer-only inference of RNNs on edge AI computing hardware with no floating-point units, so we constrained the new LSTM cell to include only integer operations. 

We propose a fully integer LSTM cell (iLSTM) in which its inference requires integer-only computation units, see Figure \ref{fig:quantized-lstm}. Our method can be applied to any RNN architecture, but we focus on LSTM networks which are the most commonly used RNNs. 

Our contributions can be summarized as 
\begin{itemize}
\item providing a quantization-aware piece-wise linear approximation algorithm to convert non-linear activation functions (e.g., sigmoid and tanh) to integer-only equivalents
\item introducing an integer-friendly normalization layer based on mean absolute deviation (MAD),
\item proposing integer-only attention,
\item combining these modules into an integer-only LSTM cell.
\end{itemize}
We also implement our method on a Huawei P20Pro smartphone, effectively showing $2\times$ speedup and $4\times$ memory compression, hence providing evidence that our method allows more RNN-based applications, such as ASR, to run on edge devices. 
%This paper extends the best industry paper award winner of ICPRAM \cite{Sari_iRNN_ICPRAM2022}.

\section{Quantization}
\label{sec:quantization}

Quantization is a process whereby an input set is mapped to a lower resolution discrete set, called the quantization set $\mathcal{Q}$. The mapping is performed from floating-points to integers (e.g. float32 to int8) or from a dense integer to another integer set with lower cardinality, e.g. int32 to int8. We follow the Quantization-Aware Training (QAT) scheme described in \cite{jacob2017quantization}. 
Given $x \in [x_{\min}, x_{\max}]$, we define the quantization process as
\begin{align}
q_x = \mathrm{q}(x) &= \Big\lfloor \frac{x}{S_x} \Big\rceil + Z_x \label{eq:quant}\\ 
r_x = \mathrm{r}(x) &= S_x(q_x - Z_x) \label{eq:dequant} \\
S_x = \frac{x_{\max} - x_{\min}}{2^b - 1},~~ & Z_x = \Big\lfloor \frac{ - x_{\min}}{S_x} \Big\rceil  \label{eq:scale-zero}
\end{align}
where the input is clipped between $x_{\min}$ and $x_{\max}$ beforehand; $\lfloor . \rceil$ is the round-to-nearest function; $S_x$ is the scaling factor (also known as the step-size); $b$ is the bitwidth, e.g. $b=8$ for 8-bit quantization, $b=16$ for 16-bit quantization; $Z_x$ is the zero-point corresponding to the quantized value of 0 (note that zero should always be included in $[x_{\min}, x_{\max}]$); $\mathrm{q}(x)$ quantizes $x$ to an integer number and $\mathrm{r}(x)$ gives the floating-point value that $\mathrm{q}(x)$ represents, i.e. $\mathrm{r}(x) \approx x$. We refer to $\{x_{\min}, x_{\max}, b, S_x, Z_x\}$ as \textbf{quantization parameters} of $x$.  Note that for inference, $S_x$ is expressed as a fixed-point integer number rather than a floating-point number, allowing for integer-only arithmetic computations \cite{jacob2017quantization}.

The following example showcases how to apply Equations (\ref{eq:quant}-\ref{eq:scale-zero}). Given a number within the previously defined range i.e. $x \in [-1, 1]$ and a target bit-width of $b=8$, $S_x$ and $Z_x$ are computed as; 
\begin{align*}
S_x &= \frac{x_{\max} - x_{\min}}{2^b - 1} = \frac{2}{ 255 } \approx 0.0078 \label{eq:appendix-scale-8}\\
Z_x &= \Big\lfloor \frac{ - x_{\min}}{S_x} \Big\rceil = \Big\lfloor \frac{1}{S_x} \Big\rceil = 128. %\label{eq:appendix-zp-8}
\end{align*}

Assuming $x=0.2$,
\begin{align*}
\mathrm{q}(x) &= \Big\lfloor \frac{x}{S_x} \Big\rceil + Z_x =  \Big\lfloor \frac{0.2}{0.0078} \Big\rceil + 128 = 154,
\end{align*}
hence $q_x=154$. The corresponding floating-point representation of $q_x$, $r_x$, is
\begin{align*}
    \mathrm{r}(x) &= S_x(q_x - Z_x) \\ 
    &= 0.0078(154 - 128) = 0.2028.
\end{align*}
Therefore, after quantizing the floating-point value $x=0.2$, its dequantized value $r_x$ is $0.2028$, so the error introduced is $\epsilon = \mid x-r_x \mid=0.0028$. The purpose of QAT is to try to force the network to be insensitive to noise introduced by these $\epsilon$ errors. The error introduced by quantization is bounded by  $\epsilon \leq \frac{S_x}{2}$. The smaller the scaling factor, the better the precision. The solutions to make the scaling factors small are either to have a smaller range $[x_{\min}, x_{\max}]$ or to use a bigger bitwidth $b$. In practice, for a given bitwidth, we would like to have a large range, including all possible values the network can produce while reducing the error as much as possible. 

\subsection{Integer-only arithmetic}
\label{sec:quant-operators}

In order to map floating-point operations to integer-only operators, all element-wise multiplication and additions must be adapted to use Equations (\ref{eq:quant}-\ref{eq:scale-zero}). We describe this mapping below. \cite{jacob2017quantization} describes how to do this for matrix-multiplications and other operators. 

At inference, the inputs to any operation are already quantized. We define two quantized inputs; $q_a$ and $q_b$ and a quantized output, $q_c$, along with their respective scaling factors $S_a$, $S_b$, $S_c$ and zero-points $Z_a$, $Z_b$, $Z_c$. 

The multiplication operation does not require $q_a$ and $q_b$ to share quantization parameters therefore, it is defined as

{\small\begin{equation}
    q_c = \Big\lfloor \frac{S_a S_b}{S_c}\Big(q_a q_b - q_a Z_b - q_b Z_a + Z_a Z_b \Big) \Big\rceil + Z_c
\label{eq:quantmul}
\end{equation}
}
where the multiplication $q_a q_b$ is a $b$-bit multiplication (example given in Appendix \ref{appendix:qmul}). 

Addition can take two forms based on the quantization parameters of $q_a$ and $q_b$. If they share the same quantization parameters (i.e. $S_b=S_a$ and $Z_b=Z_a$), it can be computed as follows,
\begin{align}
    q_c = \Big\lfloor \frac{S_a}{S_c}(q_a + q_b - 2Z_a) \Big\rceil + Z_c \label{eq:qadd-same}
\end{align}
where $q_a + q_b$ is a $b$-bit addition. However, if they do not share the same quantization parameters, 
\begin{align}
q_c = \Big\lfloor \frac{S_a}{S_c}(q_a - Z_a) + \frac{S_b}{S_c}(q_b - Z_b)\Big\rceil + Z_c \label{eq:qadd-diff}
\end{align}
rescaling $q_a$ and $q_b$ is needed, which means the addition has to be computed with more bitwidth, e.g. 32-bit (examples given in Appendix \ref{sec:appendix-qadd}). 

\subsection{Inference}
\label{sec:appendix-integer-only}
For fully-integer inference, all variables must be expressed as integers. However, the scaling-factor $S_x$ described in Equation (\ref{eq:scale-zero}) is a floating-point value. To solve this, $S_x$ can be represented as fixed-point integers, computed off-line (after training but before inference). This enables the inference to run on edge devices which typically have limited power and data format support \cite{courville2019deep}. 

A fixed-point number is an integer representation of a floating-point number, and is described using what is known as Q-format. A fixed-point integer having format $Q_{i.f}$ is said to have one sign bit, $i$ integral bits and $f$ fractional bits, bitwidth $1+i+f$, a resolution of $2^{-f}$ and span the range $[-2^{-f},2^i-2^{-f}]$. For instance, $Q_{3.4}$ represents numbers in $[-8, 7.9375]$ with resolution $0.0625$ and $Q_{0.30}$ represents numbers in $[-1, 0.99999]$ with $\approx 9\times10^{-10}$. Assuming we do not have to worry about overflows, quantizing a floating-point number $M$ to a fixed-point integer $M_{fx}$ is given by,
\begin{align}
    M_{fx} = \lfloor 2^f M\rceil. \label{eq:float-to-fx}
\end{align}
Converting a fixed-point integer $M_{fx}$ to a floating-point number $M$ is given by,
\begin{align}
    M = 2^{-f}M_{fx} \label{eq:fx-to-float}
\end{align}
An equivalent expression to (\ref{eq:fx-to-float}) is 
\begin{align}
    M = M_{fx} \gg f
\end{align}
where $\gg$ is the right shift operator. This can be incorporated into floating-point  expressions, often of the form $q_y = \lfloor Mq\rceil + Z_y$, which can be expressed as
\begin{align}
    q_y &= \Big\lfloor 2^{-f}M_{fx}q\Big\rceil + Z_y \\
    &= M_{fx}q \gg f + (M_{fx}q \gg (f-1))~\&~0\textrm{x}1 + Z_y.
\end{align}

Note that $\&~0\textrm{x}1$ masks every bit in of the number except the rightmost one, e.g. $2\&~0\textrm{x}1 = 0$ because 2 is $(10)_2$ but $3\&~0\textrm{x}1 = 1$ because 3 is $(11)_2$. It returns 0 if your number is even and 1 if your number is odd. We extracted the negative sign out of the binary shifts, as shifting on negative number has different behaviour depending on the hardware and software implementation. Note that in practice, when we know a constant can only be positive, it will be expressed as an unsigned fixed-point integer and therefore the sign bit can be allocated to either the integral or fractional part. More details on fixed-point arithmetic are provided in Appendix \ref{sec:appendix-fixed-point}.

\section{Recurrent Neural Network}

\subsection{Related Work}

In recent years, neural network quantization research \cite{jacob2017quantization,courbariaux2016bnn,darabi2018bnnplus, esser2020lsq} has enabled exploration into the quantization of RNN architectures. \cite{ott2016rnnlimitprec}, which explores low-bit quantization of weights for RNNs, showed that binarizing weights leads to a massive accuracy drop, but ternarizing them keeps the model performance. \cite{courbariaux2016bnn} demonstrates that quantizing RNNs to extremely low bits is challenging; they quantize weights and matrix product to 4-bit, but other operations such activation functions are computed in full-precision. 

Low-precision RNNs has also been a recent focal point in industry research. \cite{he2016effectivernn} introduces Bit-RNN and improves 1-bit and 2-bit RNNs quantization by constraining values within a fixed range carefully; they keep activation computation and element-wise operations in full-precision. \cite{kapur2017lowprecrnn} builds upon Bit-RNN and propose a low-bit RNN with minimal performance drop, but they increase the number of neurons to compensate for performance drop; they run activation functions in floating-point precision as well. \cite{hou2019normalization} quantizes LSTM weights to 1-bit and 2-bit and empirically show that low-bit quantized LSTMs suffer from exploding gradients. It has been shown that gradient explosion can be alleviated using normalization layers and leads to the successful training of low bit weights \cite{ardakani2018learning}. \cite{sari2020normalization} studied the effect of normalization in low-bit networks theoretically and proved that low-bit training without normalization operation is mathematically impossible; their work demonstrates the fundamental importance of involving normalization layers in quantized networks. 

\cite{wu2016gnmt} is a pioneering work in LSTM quantization, which demonstrates speed-up inference of large-scale LSTM models with limited performance drop by partially quantizing RNN cells. Their proposed method is tailored towards specific hardware. They use an 8-bit integer for matrix multiplications and a 16-bit integer for tanh, sigmoid, and element-wise operations but do not quantize attention.  \cite{bluche2020sonosquantlstm}  propose an effective 8-bit integer-only LSTM cell for Keyword Spotting application on microcontrollers. They enforce weights and activations to be symmetric on fixed ranges $[-4, 4]$ and $[-1,1]$. This prior assumption about the network's behavior restricts generalizing their approach for many RNN models. They propose a look-up table of 256 slots to represent the quantized tanh and sigmoid activations. However, the look-up table memory requirement explodes for a bigger bitwidth. Their solution does not serve complex tasks such as automatic speech recognition due to large look-up table memory consumption. While demonstrating strong results on the Keyword Spotting task, their assumptions on quantization range and bitwidth make their method task-specific. 

To explain the methodology in the following subsections, we use the common linear algebra notation and use plain symbols to denote scalar values, e.g. $x\in \mathbb R$, bold lower-case letters to denote vectors, e.g. $\textbf{x} \in \mathbb{R}^{n}$, and bold upper-case letters to denote  matrices, e.g. $\textbf{X} \in \mathbb{R}^{m \times n}$. The element-wise multiplication is represented by $\odot$. 
\subsection{Long short term memory}
We define an LSTM cell as
\begin{align}
\begin{pmatrix} 
\mathbf{i}_t \\ 
\mathbf{f}_t \\ 
\mathbf{j}_t \\ 
\mathbf{o}_t 
\end{pmatrix} &= \mathbf{W}_x\mathbf{x}_t + \mathbf{W}_h\mathbf{h}_{t-1}, \label{eq:gates}\\
\mathbf{c}_t &= \sigma(\mathbf{f}_t) \odot \mathbf{c}_{t-1} + \sigma(\mathbf{i}_t) \odot \text{tanh}(\mathbf{j}_t), \label{eq:lstm-cell-state}\\
\mathbf{h}_t &= \sigma(\mathbf{o}_t) \odot \text{tanh}(\mathbf{c}_t),
\label{eq:vanilla-lstm}
\end{align}
where $\sigma(\cdot)$ is the sigmoid function; $n$ is the input hidden units dimension, and $m$ is the state hidden units dimension; $\mathbf{x}_t \in \mathbb{R}^{n}$ is the input for the current timestep $t \in \{1, ..., T\}$; $\mathbf{h}_{t-1} \in \mathbb{R}^{m}$ is the hidden state from the previous timestep and $\mathbf{h}_0$ is initialized with zeros; $\mathbf{W}_x \in \mathbb{R}^{4m \times n}$ is the input to state weight matrix; $\mathbf{W}_h \in \mathbb{R}^{4m \times m}$ is the  state to state weight matrix; $\{\mathbf{i}_t, \mathbf{f}_t, \mathbf{o}_t\} \in \mathbb{R}^{m}$ are the pre-activations to the \{input, forget, output\} gates; $\mathbf{j}_t \in \mathbb{R}^{m}$ is the pre-activation to the cell candidate; $\{\mathbf{c}_t, \mathbf{h}_t\} \in \mathbb{R}^{m}$ are the cell state and the hidden state for the current timestep, respectively. We omit the biases for the sake of notation simplicity. 
For a bidirectional LSTM (BiLSTM) the output hidden state at timestep $t$  is the concatenation of the forward  hidden state $\overrightarrow{\mathbf{h}}_t$ and the backward  hidden state $\overleftarrow{\mathbf{h}}_t$, $[\overrightarrow{\mathbf{h}}_t; \overleftarrow{\mathbf{h}}_t]$.

\subsection{LayerNorm}
\label{sec:layernorm}
Layer normalization \cite{ba2016layernorm} standardizes inputs across the hidden units dimension   with zero location and unit scale. Given hidden units $\mathbf{x} \in \mathbb{R}^H$, LayerNorm is defined as 

\begin{eqnarray}
\mu = \frac{1}{H}\sum_{i=1}^H x_i \quad\quad  \hat{x}_i = x_i - \mu \label{eq:centered-input} \\
\std^2 = \frac{1}{H}\sum_{i=1}^H\hat{x}_i^2 \quad\quad  \std = \sqrt{\std^2} \label{eq:std}\\ 
\mathrm{LN}(\mathbf{x})_i = y_i = \frac{\hat{x_i}}{\std} \label{eq:normalization}
\end{eqnarray}

\begin{equation}
\forall i,j 1\leq i,j\leq n 
   \quad\text{and}\quad 
c_{ij} = \sum a_{ik} b_{kj}
\end{equation}

where $\mu$ (\ref{eq:centered-input}) is the hidden unit mean, $\hat{x}_i$ (\ref{eq:centered-input}) is the centred hidden unit $x_i$, $\std^2$ (\ref{eq:std}) is the hidden unit variance, and $y_i$ (\ref{eq:normalization}) is the normalized hidden unit. In practice, one can scale $y_i$ by a learnable parameter $\gamma$ or shift by a learnable parameter $\beta$. The LayerNormLSTM cell is defined as  in \cite{ba2016layernorm}.

\subsection{Attention}
Attention is often used in encoder-decoder RNN architectures \cite{bahdanau2015attention, chorowski2015speechattention, wu2016gnmt}. We employ Bahdanau attention, also called additive attention \cite{bahdanau2015attention}. The attention mechanism allows the decoder network to attend to the variable-length output states from the encoder based on their relevance to the current decoder timestep. At each of its timesteps, the decoder extracts information from the encoder's states and summarizes it as a context vector, 
\begin{align}
\mathbf{s}_t &= \sum_{i=1}^{T_{\mathrm{enc}}} \alpha_{ti}\odot\mathbf{h}_{\mathrm{enc}_i} \\
\alpha_{ti} &= \frac{\text{exp}(e_{ti})}{\sum_{j=1}^{T_{\mathrm{enc}}}\text{exp}(e_{tj})} \label{eq:attn-alpha} \\
e_{ti} &= \mathbf{v}^\top \text{tanh}(\mathbf{W}_q\mathbf{h}_{t-1} + \mathbf{W}_k\mathbf{h}_{enc_i}) \label{eq:attn-qk}
\end{align}
where $\mathbf{s}_t$ is the context at decoder timestep $t$ which is a weighted sum of the encoder hidden states outputs $\mathbf{h}_{\mathrm{enc}_i} \in \mathbb{R}^{m_{\mathrm{enc}}}$ along encoder timesteps $i \in \{1, ..., T_{\mathrm{enc}}\}$ 
; $0<\alpha_{ti}<1$ are the attention weights attributed to each encoder hidden states based on the alignments $e_{ti} \in \mathbb{R}$; $m_{\mathrm{dec}}$ and $m_{\mathrm{enc}}$ are respectively the decoder and encoder hidden state dimension; 
$\{\mathbf{W}_q \in \mathbb{R}^{m_{\mathrm{att}} \times m_{\mathrm{dec}}}, \mathbf{W}_k \in \mathbb{R}^{m_{\mathrm{att}} \times m_{\mathrm{enc}}} \}$ are the weights matrices of output dimension $m_{\mathrm{att}}$ respectively applied to the query $\mathbf{h}_{t-1}$ and the keys $\mathbf{h}_{\mathrm{enc}_i}$; $\mathbf{v} \in \mathbb{R}^{m_{\mathrm{att}}}$ is a learned weight vector. The context vector is incorporated into the LSTM cell by modifying (\ref{eq:gates}) to 
\begin{eqnarray}
\begin{pmatrix} 
\mathbf{i} \\ 
\mathbf{f} \\ 
\mathbf{j} \\ 
\mathbf{o} 
\end{pmatrix} = \mathbf{W}_x\mathbf{x}_t + \mathbf{W}_h\mathbf{h}_{t-1} + \mathbf{W}_{s}\mathbf{s}_t
\end{eqnarray}
where $\mathbf{W}_{s} \in \mathbb{R}^{4m_{\mathrm{dec}} \times m_{\mathrm{enc}}}$.

\section{Methodology}
In this section, we describe our task-agnostic quantization-aware training method to enable integer-only RNN (iRNN). 

\subsection{Integer-only activation}
\label{sec:quant-act}

First, we need to compute activation functions without relying on floating-point operations to take the early step towards an integer-only RNN. At inference, the non-linear activation is applied to the quantized input $q_x$, operations using integer-only arithmetic and output the quantized result $q_y$. Clearly, given the activation function $f$, $q_y = q(f(q_x))$;
as the input and the activation and output are both quantized, we obtain a discrete mapping from $q_x$ to $q_y$. There are several ways to formalize this operation. The first  solution is a Look-Up Table (LUT), where $q_x$ is the index and $q_y = \text{LUT}[q_x]$. For 8-bit quantization, a LUT would require 256 bytes of storage ($2^8$ 8-bit elements), however 16-bit quantization would require up to 128 kilobytes of memory ($2^{16}$ 16-bit elements). Therefore, this method does not scale well and is not cache-friendly for large indexing bitwidths. The second solution approximates the full-precision activation function using a fixed-point integer Taylor approximation, but the amount of computations grows as the approximation order grows. 

We propose to replace all non-linear activation functions of an LSTM cell with integer-only arithmetic by approximating them using a \textbf{Quantization-Aware PWL} function which defines the PWL pieces at training time. This enables the precision of the PWL to be adapted to the task at hand and provides highly accurate data-dependent activation approximation with fewer pieces.

\begin{figure}
\centering
\includegraphics[width=0.45\linewidth]{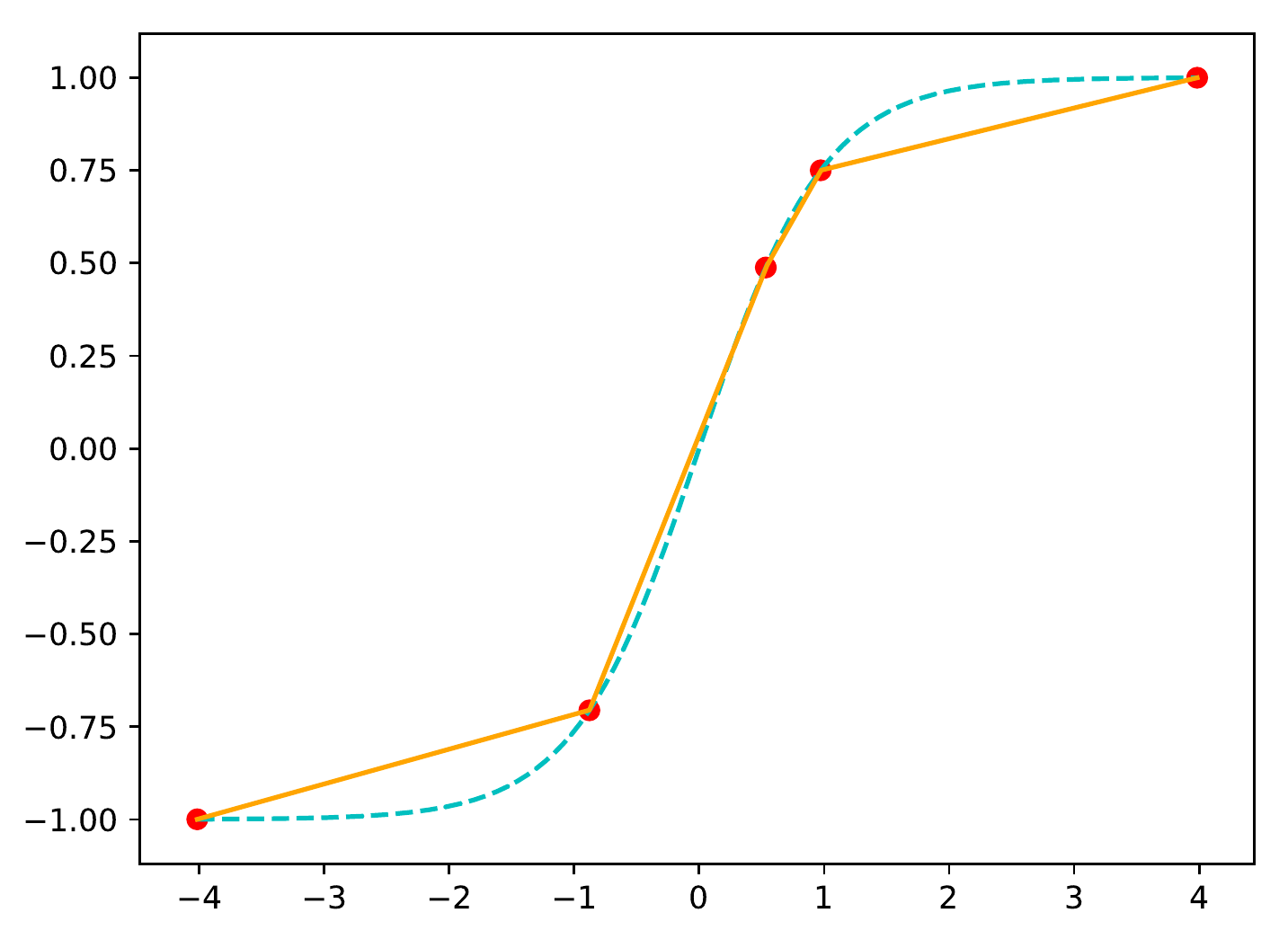}
\includegraphics[width=0.45\linewidth]{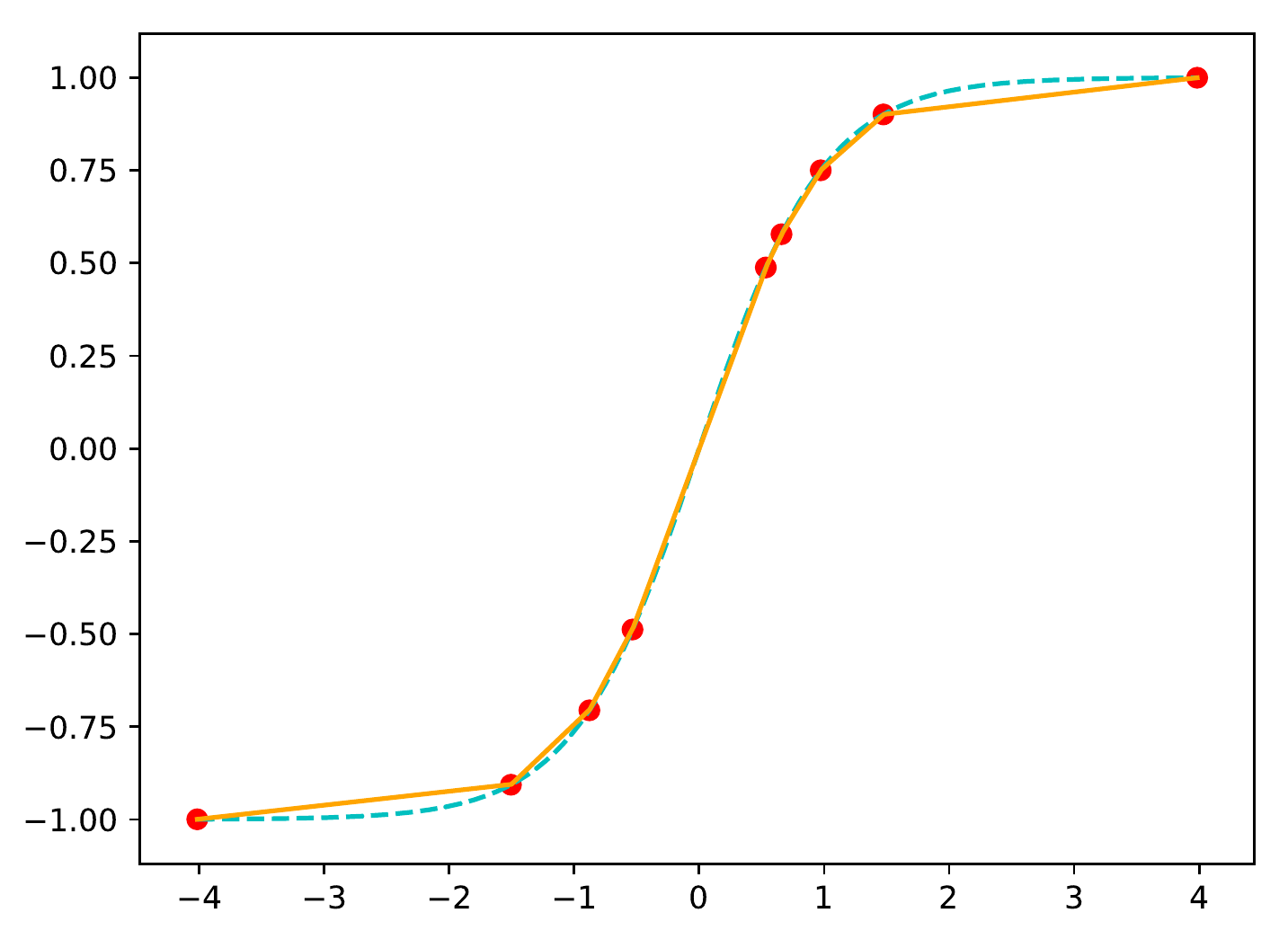}\\
 \includegraphics[width=0.45\linewidth]{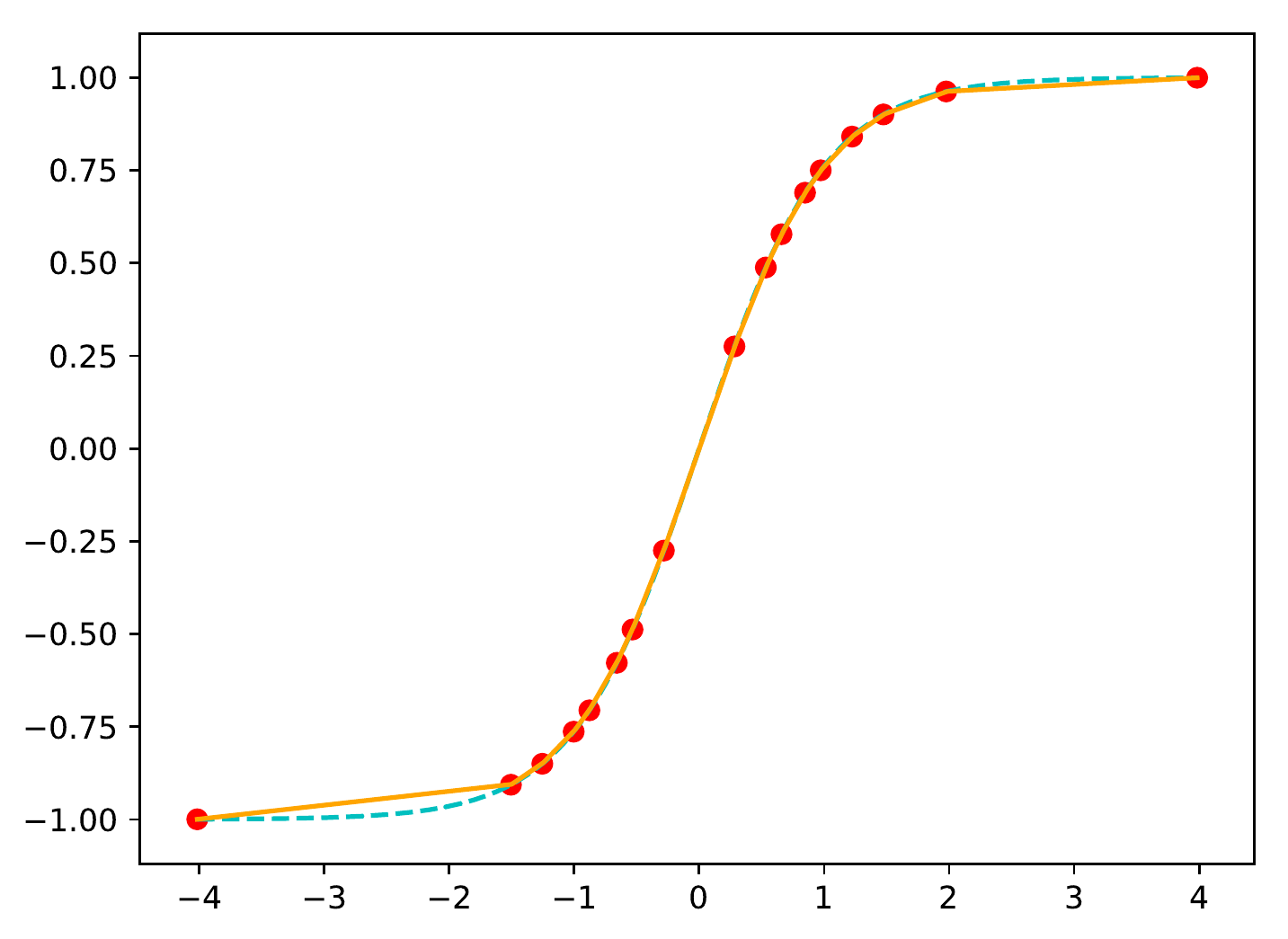}
 \includegraphics[width=0.45\linewidth]{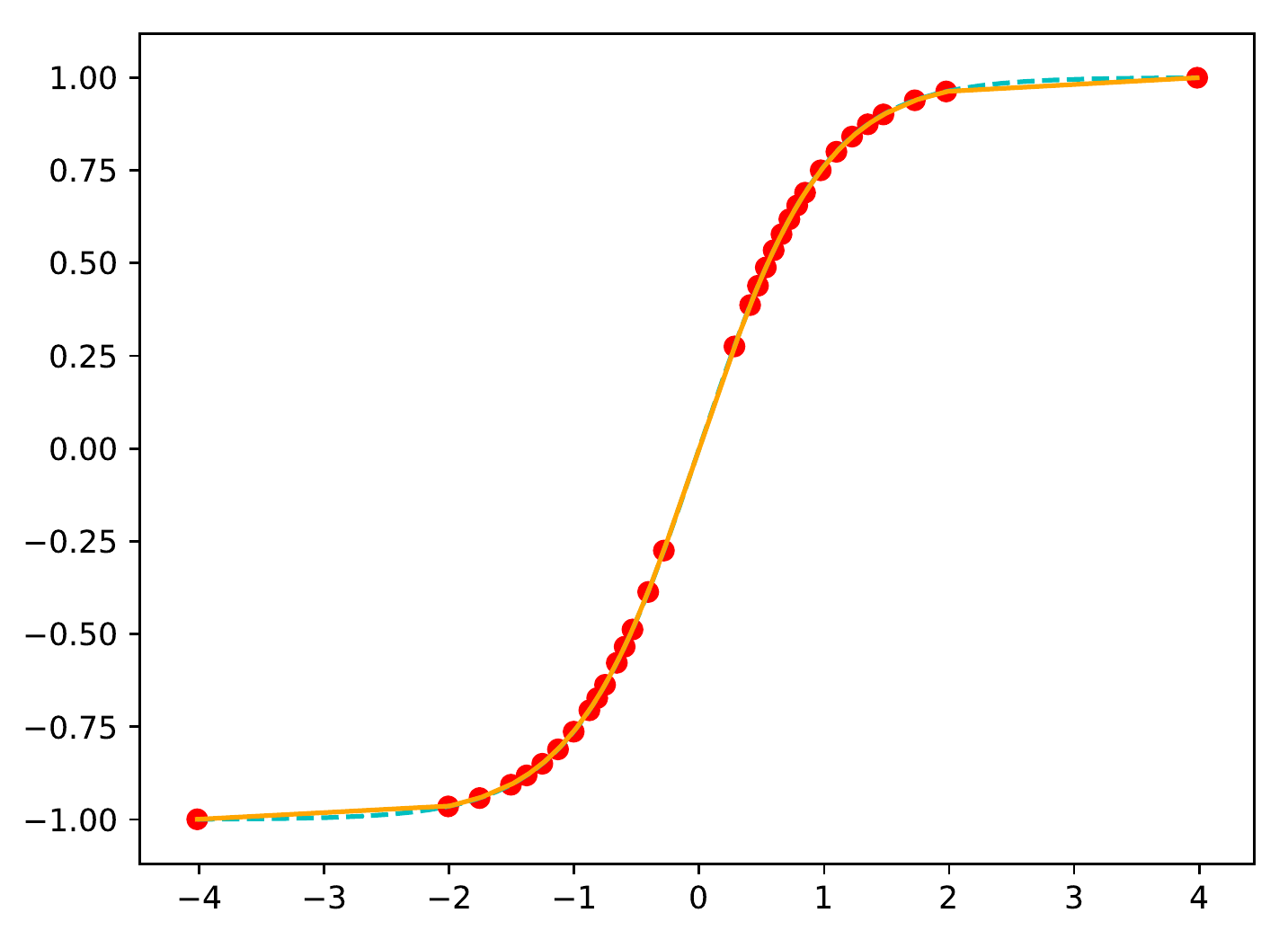}
\caption{
$\tanh$ approximations using quantization-aware PWLs with 4 pieces (left panel), 16 pieces (right panel) using (\ref{eq:pwl}). The dashed cyan curves are the true tanh functions, while the solid orange curves are its approximation from Algorithm \ref{alg:pwl}. Red dots are the knots. The more we add pieces, the better the approximation is. Our algorithm can prioritize sections of the function with more curvatures; see Appendix \ref{sec:appendix-A} Figure \ref{fig:other-pwl} for more instances.}
\label{fig:tanh-pwl}
\end{figure}

A PWL is defined as follows,
\begin{eqnarray}
g(x) = \sum_{i=1}^{N} \mathbbm{1}_{[k_i, k_{i+1})} \Big(a_i(x - k_i) + b_i \Big), \label{eq:pwl}
\end{eqnarray}
where $N$ is the number of linear pieces defined by $N + 1$ knots (also known as cutpoints or breakpoints); $\{a_i, k_i, b_i=f(k_i)\}$ are the slope, the knot, and the intercept of the $i^{\text{th}}$ piece respectively;  $\mathbbm{1}_{A}(x) = 1$ is the indicator function on $A$.  The more the linear pieces, the better the activation approximation is (see Figure \ref{fig:tanh-pwl}). A PWL is suitable for simple fixed-point integer operations. It only relies on basic arithmetic operations and is easy to parallelize because the computation of each piece is independent. Therefore, the challenge is to select the knot locations that provide the best PWL approximation to the original function $f$. Note in this regime, we only approximate the activation function on the subset corresponding quantized inputs and not the whole full-precision range. In our proposed method if $x=k_i$ then $g(x) = g(k_i) = b_i$, i.e. recovers the exact output $f(k_i)$. Hence, if the PWL has $2^b$ knots (i.e. $2^b - 1$ pieces), it is equivalent to a look-up table representing the quantized activation function. Thus, we constraint the knots to be a subset of the quantized inputs of the function we are approximating (i.e. $\{k_i\}_{i=1}^{N+1} \subseteq \mathcal{Q}$).

\begin{algorithm}
\caption{Piece-wise linear approximation of activation $f$}\label{alg:pwl}
\includegraphics[width=0.9\textwidth]{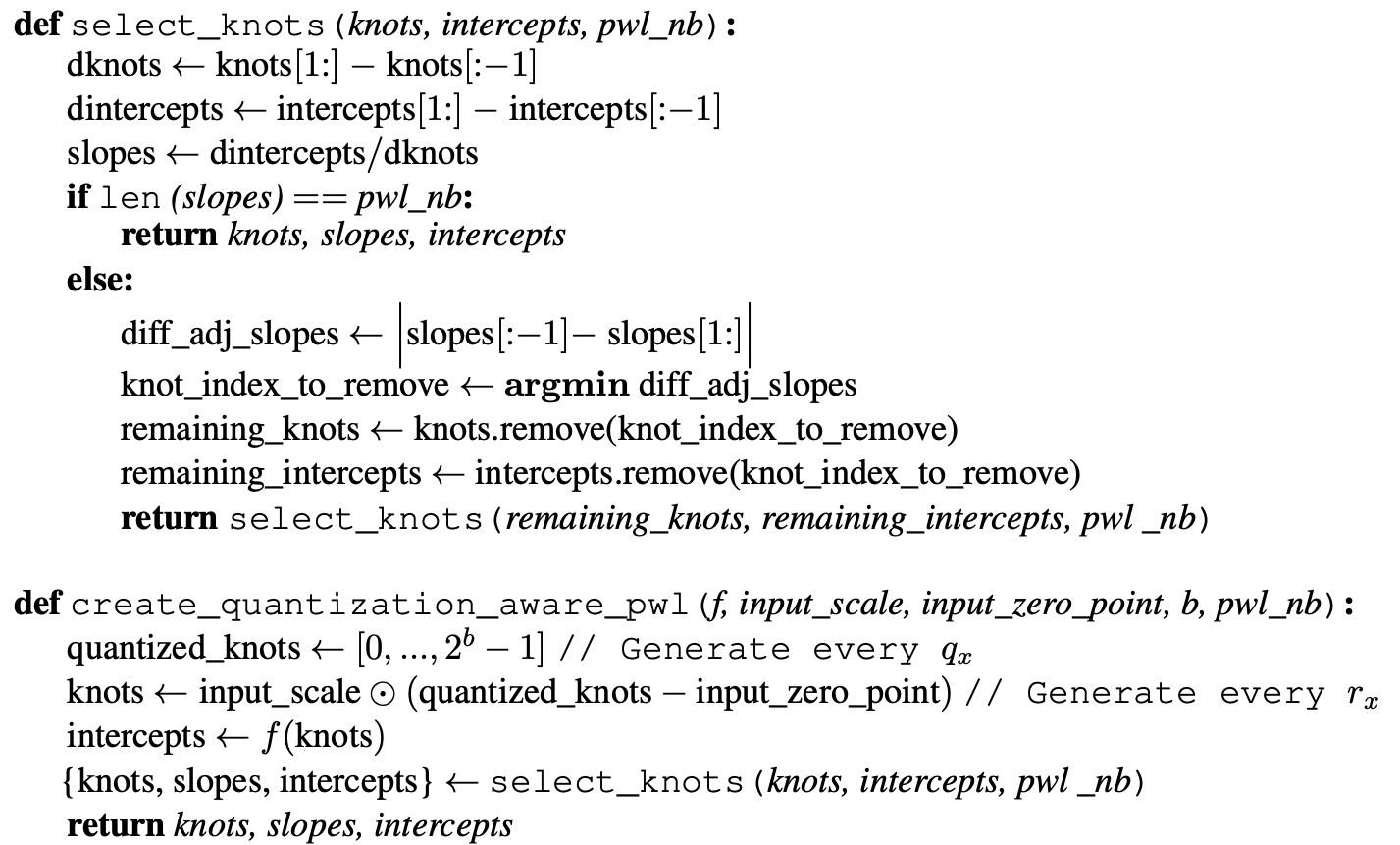}
\end{algorithm}

We propose a recursive greedy algorithm, Algorithm \ref{alg:pwl}, to locate the knots during the quantization-aware PWL. The algorithm starts with $2^b - 1$ pieces and recursively removes one knot at a time until it reaches the specified number of pieces. The absolute differences between adjacent slopes are computed, and the shared knot from the pair of slopes that minimizes the absolute difference is removed; see Appendix \ref{sec:appendix-A}, Figure \ref{fig:pwl-algo}. Applied only once per training step, the generic nature of this algorithm enables it to be applied to any nonlinear function. Note that the PWL is specific to a given set of quantization parameters, i.e. the quantization parameters are kept frozen after its creation.

At inference, the quantization-aware PWL is computed as follows
\begin{eqnarray}
q_y = \Big\lfloor \sum_{i=1}^{N} \mathbbm{1}_{[q_{k_i}, q_{k_{i+1}})} \Big(\frac{S_x a_i}{S_y}(q_x - q_{k_i}) + \frac{b_i}{S_y}\Big) \Big\rceil  + Z_y,
\end{eqnarray}
where the constants are expressed as fixed-point integers.

 \subsection{Integer-only normalization}
 \label{sec:madnorm}
 Normalization greatly helps the convergence of quantized networks \cite{hou2019normalization, sari2020normalization}. There are many location and scale measures to define a normalization operation. The commonly used measure of dispersion is the standard deviation to define normalization, which is imprecise and costly to compute on integer-only hardware. However, the mean absolute deviation (MAD) is integer-friendly and defined as 
\begin{equation}
d = \frac{1}{H}\sum_{i=1}^{H} \abs{ x_i - \mu }
\end{equation}
 While the mean minimizes the standard deviation, the median minimizes MAD. We suggest measuring deviation from the mean for two reasons: i) the median is computationally more expensive ii) the absolute deviation from the mean is closer to the standard deviation. For Gaussian data, the MAD is $\approx 0.8 \std$ so that it might be exchanged with standard deviation. We propose to LayerNorm in LSTM with MAD instead of standard deviation and refer to it as MadNorm., where \eqref{eq:normalization} is replaced by
 \begin{equation}
 y_i = \frac{\hat{x}_i}{d}.
 \label{eq:mad-normalization}
 \end{equation}
 MadNorm involves simpler operations, as there is no need to square and no need to take the square root while taking absolute value instead of these two operations is much cheaper. The values $\{\mu, \hat{x}_i, d, y_i\}$ are 8-bit quantized and computed as follows
\begin{align}
q_{\mu} &= \Big\lfloor \frac{S_x}{S_\mu N}\Big(\sum_{i=1}^Nq_{x_i} - NZ_x\Big) \Big\rceil + Z_\mu,
\label{eq:qmean}\\
 q_{\hat{x}_i} &=\Big\lfloor \frac{S_x}{S_{\hat{x}}}(q_{x_i} - Z_x) - \frac{S_\mu}{S_{\hat{x}}}(q_{\mu} - Z_\mu) \Big\rceil + Z_{\hat{x}},
 \label{eq:qcentered}\\
 q_d &= \Big\lfloor \frac{S_{\hat{x}}}{S_{d}N}\sum_{i=1}^N \abs{ q_{\hat{x}_i} -  Z_{\hat{x}}}  \Big\rceil + Z_{d},
 \label{eq:qmad}\\
 q_{y_i} &= \Big\lfloor \frac{\frac{S_{\hat{x}}}{S_y S_d} (q_{\hat{x_i}} - Z_{\hat{x}})}{\mathrm{\max}(q_d, 1)} \Big\rceil + Z_y,
\label{eq:qmadnorm}
\end{align}
where all floating-point constants can be expressed as fixed-point integer numbers, allowing for integer-only arithmetic computations. Note that (\ref{eq:qmean}-\ref{eq:qmadnorm}) are only examples of ways to perform integer-only arithmetic for MadNorm, and may change depending on the software implementation and the target hardware.
We propose to quantize $\{\mathbf{v}, \mathbf{W}_q, \mathbf{W}_k\}$ to 8-bit. The vectors $\mathbf{h}_{t-1}$ and $\mathbf{h}_{\mathrm{enc}_i}$ are quantized thanks to the previous timestep and/or layer. The matrix multiplications in (\ref{eq:attn-qk}) are performed in 8-bit, and their results are quantized to 8-bit, each with its quantization parameters. Since those matrix multiplications do not share the same quantization parameters, the sum  (\ref{eq:attn-qk}) requires proper rescaling, and the result is quantized to 16-bit. We found that 8-bit quantization adds too much noise, thus preventing the encoder-decoder model from working correctly. The tanh function in (\ref{eq:attn-qk}) is computed using quantization-aware PWL, and its outputs are quantized to 8-bit. The alignments $e_{ti}$ are quantized to 16-bit (\ref{eq:attn-qk}). The exponential function in $\alpha_{ti}$ is computed using a quantization-aware PWL, with its outputs quantized to 8-bit. We found that quantizing the softmax denominator (\ref{eq:attn-alpha}) to 8-bit introduces too much noise and destroys attention. 8-bit attention does not offer enough flexibility and prevents fine-grained decoder attention to the encoder. We left the denominator in a 32-bit integer value. The context vector $\mathbf{s}_t$ is quantized to 8-bit. Note that in practice we shift the inputs of softmax for numerical stability reasons  $e_{ti}\leftarrow e_{ti} - \max\limits_j e_{tj}$.

\begin{theorem}[Scale convergence]
\label{theo:scale}
Suppose $\{ X_i \}_{i=1}^{\infty}$ is a sequence of integrable pairwise independent random variables defined on the same probability space $(\Omega, \mathcal F, P)$ with $\mu=\E(X_i)$. Then $D_n={1\over n}\sum_{i=1}^n \abs{ X_i-\mu} $ converges almost surely to $\tilde \sigma = \E(\abs{ X-\mu} ).$
\end{theorem}
{\bf Proof:} Integrability assumption implies existence of $\mu=\E(X_i)<\E(\abs{ X_i} ) <\infty$ and hence existence of $\E\abs{ X_i-\mu} <\E\abs{ X_i} +\abs{\mu} <\infty$. The desired result then follows from the Strong Law of Large Numbers to $Y_i= \abs{ X_i-\mu}. \ \blacksquare$

One may prove the central limit theorem by replacing absolute integrability with square integrability, and exchanging pairwise independence with mutual independence. 
Convergence to the population scale $\tilde \sigma$ in Theorem~\ref{theo:scale} paves the way to show that our MadNorm enjoys a concentration inequality similar to LayerNorm.

\begin{theorem}[Concentration inequality]
\label{theo:con}
Suppose $X$ defined on $(\Omega, \mathcal F, P)$ is an integrable random variable with mean $\mu$. Then for a positive $k$, 

\[ P\left(\abs{X-\mu \over \tilde \sigma}<k\right)\geq 1-{1\over k}
\]

\end{theorem}
{\bf Proof:} 
The result follows from Markov's inequality, see \cite{Sari_iRNN_ICPRAM2022} for details. 
%Take $Y=\abs{X-\mu \over \tilde \sigma} $. The random variable $X$ is absolutely integrable, and so is $Y$. 
%\begin{eqnarray*}
%\E(Y) &=& \int_0^\infty Y dP = \int_0^k YdP + \int_k^\infty YdP\\
%&\geq& 0+ k\int_k^\infty YdP = k\Pr(Y>k), 
%\end{eqnarray*}
%it follows immediately that $\Pr\left( \abs{X-\mu \over \tilde \sigma}>k\right)\leq {1\over k}. 
$ \blacksquare$\\

Theorem~\ref{theo:con} assures that independent of the distribution of data, this MadNorm brings the mass of the distribution around the origin. This is somehow expected from any normalization method. Therefore, it is not surprising that LayerNorm also has a similar property. Therefore, LayerNorm and MadNorm assure that the tail probability far from the origin is negligible. 

There is a slight difference between the concentration inequality of LayerNorm and MadNorm. The LayerNorm provides a tighter bound, i.e., the bound in Theorem~\ref{theo:con} changes from $1-{1\over k}$ to $1-{1\over k^2}$, but it also requires more assumptions like the square integrability of $X$.

\subsection{Integer-only attention}
\label{sec:qattention}
Attention plays a crucial role in modern encoder-decoder architectures. The decoder relies on attention to extract information from the encoder and provide predictions. Attention is the bridge between the encoder and the decoder. Careless quantization of attention breaks apart the decoder due to quantization noise. 

\subsection{Integer-only LSTM }
\label{sec:qlstm}
A standard LSTM cell comprises  matrix multiplications, element-wise additions, element-wise multiplications, tanh, and sigmoid activations (\ref{eq:gates} - \ref{eq:vanilla-lstm}). We quantize the weights matrices $\mathbf{W}_x$ and $\mathbf{W}_h$ to 8-bit. The inputs $\mathbf{x}_t$ and hidden states $\mathbf{h}_{t-1}$ are already 8-bit quantized from the previous layer and from the previous timestep. The cell states $\mathbf{c}_t$ are theoretically unbounded (\ref{eq:lstm-cell-state}); therefore the amount of quantization noise potentially destroys the information carried by $\mathbf{c}_t$, if it spans a large range. When performing QAT on some pre-trained models, it is advised to quantize $\mathbf{c}_t$ to 16-bit. Therefore, $\mathbf{c}_t$  is 8-bit quantized unless stated otherwise but can be quantized to 16-bit if necessary. Matrix multiplications in (\ref{eq:gates}) are performed with 8-bit arithmetic, and their outputs are quantized to 8-bit based on their respective quantization parameters. The sum between the two matrix multiplications outputs in (\ref{eq:gates}) requires proper rescaling because they do not share the same quantization parameters.

The sum results are quantized to 8-bit; however, 16-bit quantization might be necessary for complex tasks. The sigmoid and tanh activations in  (\ref{eq:lstm-cell-state}) and (\ref{eq:vanilla-lstm}) are replaced with their own quantization-aware PWL, and their output is always quantized to 8-bit. The element-wise multiplications operations are distributive, and sharing quantization parameters is not required. In (\ref{eq:lstm-cell-state}), the element-wise multiplications are quantized to 8-bit, but can be quantized to 16-bit if $\mathbf{c}_t$ is quantized to 16-bit as well; the element-wise additions are quantized based on $\mathbf{c}_t$'s bitwidth (i.e. 8-bit or 16-bit).

The element-wise multiplications between sigmoid and tanh in (\ref{eq:vanilla-lstm}) are always quantized to 8-bit because $\mathbf{h}_t$ are always quantized to 8-bit. We obtain an integer-only arithmetic LSTM cell following this recipe, see Figure. \ref{fig:quantized-lstm}. For LSTM cells with LayerNorm quantized MadNorm layers are used instead of LayerNorm. Appendix \ref{sec:appendix-lstm-continue} 
provides details about the quantization of other types of layers in an LSTM model. 

\section{Experimental results}
\label{sec:experiments}
We evaluate our proposed method, iRNN, on language modeling and automatic speech recognition. We also implemented our approach on a smartphone to benchmark inference speedup, see Section \ref{sec:inference}.
\subsection{Language modeling on PTB}
\begin{table}
{\small
\begin{center}
\begin{tabular}{ l | c | c }
      \hline 
	  LayerNorm LSTM  & val & test\\
	  \hline
	  Full-precision & $98.58 \pm 0.35$ & $ 94.84 \pm 0.21$ \\
	  \hline
	  \hline

      PWL4 & $101.40 \pm 0.70$ & $98.11 \pm 0.75$ \\ 
      PWL8 & $98.14 \pm 0.11$ & $95.03 \pm 0.16$ \\
      PWL16 & $98.09 \pm 0.06$ & $94.92 \pm 0.05$ \\ 
      PWL32 & $\mathbf{97.97} \pm 0.01$ & $\mathbf{94.81} \pm 0.02$ \\ 
\end{tabular} 
%}
\end{center}
}
\caption{Word-level perplexities on PTB with a LayerNorm LSTM and quantized models with a different number of PWL pieces. LayerNorm is replaced with MadNorm for the quantized models (iRNN). Best results are averaged across 3 runs $\pm$ standard deviation.}
\label{table:ptb}
\end{table}
As a proof of concept, we perform several experiments on full-precision and entirely 8-bit quantized models on the Penn TreeBank (PTB) dataset \cite{marcus1993ptb}. We report perplexity per word as a performance metric.
For the quantized models, the LayerNorm is replaced with MadNorm. We do not train full-precision models with MadNorm to make our method comparable with common full-precision architectures. The quantized PWL models are initialized from the best full-precision checkpoint and train for another 100 epochs. Experimental setup details are provided in Appendix \ref{sec:appendix-ptb}. We can draw two conclusions from the results presented in Table \ref{table:ptb}, i) replacing LayerNorm with MadNorm does not destroy model performance, ii) using eight linear pieces is enough to retain the performance of the model, but adding more linear pieces improves the performance. Furthermore, we could obtain even superior results in the quantized model compared to the full-precision model because of the regularization introduced by quantization errors.

\subsection{Language modeling on WikiText2}
\label{sec:wikitext2}
\begin{table}
{\small
\begin{center}
\begin{tabular}{ l | c | c }
    \hline
	  Mogrifier LSTM & val & test\\
	  \hline
	  Full-precision & $60.27 \pm 0.34$ & $58.02 \pm 0.34$ \\
	  \hline
	  \hline
      PWL8   & $ 60.91 \pm 0.04$ & $ 58.54\pm $ 0.07 \\
      PWL16 & $ 60.65 \pm 0.09 $ & $ 58.21 \pm  0.08$\\ 
      PWL32 & $ \mathbf{60.37} \pm 0.03$ & $ \mathbf{57.93} \pm 0.07$

    %   PWL4 Quantized LSTM & $ \pm $ & $ \pm $ \\ 
\end{tabular} 
%}
\end{center}
\caption{Word-level perplexities on WikiText2 with Mogrifier LSTM and quantized models with different number of PWL pieces. Best results are averaged across 3 runs $\pm$ standard deviations.}
\label{table:mogrifier-lstm}
}
 \end{table}

We evaluated our proposed method on the WikiText2 dataset \cite{merity2016wikitext} with a state-of-the-art RNN, Mogrifier LSTM \cite{melis2020mogrifier}. The original code\footnote{\url{https://github.com/deepmind/lamb}} was written in TensorFlow, and we reimplemented our version in PyTorch by staying as close as possible to the TensorFlow version. We follow the experimental setup from the authors\footnote{\url{https://github.com/deepmind/lamb/blob/254a0b0e330c44e00cf535f98e9538d6e735750b/lamb/experiment/mogrifier/config/c51c838b33a5+_tune_wikitext-2_35m_lstm_mos2_fm_d2_arms/trial_747/config}} as we found it critical to get similar results.
We use a two-layer Mogrifier LSTM. The setup and hyper-parameters for the experiments can be found in Appendix \ref{sec:appendix-mogrifier} 
to save some space.    We present our results averaged over 3 runs in Table \ref{table:mogrifier-lstm}. We use the best full-precision model, which scores $59.93$ perplexity to initialize the quantized models. Our method can produce an 8-bit quantized integer-only Mogrifier LSTM with similar performance to the full-precision model with only about $0.3$ perplexity increase for the quantized model with a PWL of 32 pieces and a maximum of about $0.9$ perplexity increase with several pieces as low as 8. Interestingly, a pattern emerged by doubling the number of pieces, as we get a decrease in perplexity by about $0.3$. We also perform a thorough ablation study of our method in Table \ref{table:mogrifier-ablation}. 
Surprisingly, we found that stochastic weight averaging for quantized models exhibits the same behavior as for full-precision models and improved performance thanks to regularization. While experiments on the PTB dataset demonstrated the potential of our method, these experiments on WikiText2 show that our proposed method can stay on par with state-of-the-art RNN models. This helps emphasize that our fully-integer iRNN model is capable of maintaining comparable accuracy to the full-precision model.

\subsection{ASR on LibriSpeech}
\begin{table}
{\small
\begin{center}
\begin{tabular}{ l | c | c | c}
    \hline
	  ESPRESSO LSTM & set & clean & other\\
	  \hline
	  Full-precision & dev & 2.99 & 8.77 \\
	  iRNN PWL96* & dev & 3.73 & 10.02 \\
	  \hline 
	  \hline 
	  Full-precision & test & 3.37 & 9.49 \\

      iRNN PWL96* & test & 4.11 & 10.71 \\

\end{tabular} 
\end{center}
\caption{WER\% on LibriSpeech with ESPRESSO LSTM (Encoder-Decoder LSTM with Attention) with LM shallow fusion. *(160 pieces were used for the exponential function)}
\label{tab:asr}
}\end{table}
ASR is a critical edge AI application and is challenging due to the nature of the task. Voice is diverse as the human voice may vary in pitch, accent, pronunciation style, voice volume, etc.  
While we showed our method is working for a competitive language modeling task, one can argue that ASR is a more practical and, at the same time, more difficult task for edge and IoT applications. Therefore, we experiment on an ASR task based on the setup of \cite{wang2019espresso} and their ESPRESSO framework\footnote{\url{https://github.com/freewym/espresso}}. We used an LSTM-based Attention Encoder-Decoder (ESPRESSO LSTM) trained on the strong ASR LibriSpeech dataset \cite{panayotov2015librispeech}. Experiments setup and hyper-parameters are provided in Appendix \ref{sec:appendix-asr}.
We initialize the quantized model from the pre-trained full-precision ESPRESSO LSTM. Our early experiments confirmed that quantizing the model to 8-bit would not give comparable results. After investigation, we noticed it was mainly due to two reasons, i) the cell states $\mathbf{c}_t$ had large ranges (e.g., $[-17, 15]$), and ii) the attention mechanism was not letting the decoder attend the encoder outputs accurately. Therefore, we quantize the pre-activation gates (\ref{eq:gates}), the element-wise multiplications in (\ref{eq:lstm-cell-state}) and cell states $\mathbf{c}_t$ to 16-bit. The attention is quantized following our described integer-only attention method. Everything else is quantized to 8-bit following our described method. The quantized model has a similar performance to the full-precision model, with a maximum of $1.25$ WER\% drop (Table \ref{tab:asr}). We believe allowing the model to train longer would reduce the gap.

 \subsection{Inference measurements}
\label{sec:inference}
\begin{table}
{
\begin{center}
\begin{tabular}{ c  c  c  c c }
	  LSTM & quantized & forward pass (ms) & iteration & speedup\\
	  & activation & & in second \\
	  \hline
	  Full-precision  & \xmark & 130 & 7.6 & $1.00\times$ \\
%	  \hline 
%	  \hline
	  iRNN w/  & PWL32 & 84 & 11.8 & $1.54\times$ \\
	  iRNN   & PWL8 & 61 & 14.9 & $\mathbf{1.95\times}$ \\
	  iRNN  & \xmark & 127 & 7.8 & $1.02\times$\\
\end{tabular} 
%}
\end{center}
\caption{Inference measurements on a Huawei P20 Pro smartphone, based on a custom fork from PyTorch 1.7.1. The model is one LSTM cell with a state size of 400. Milliseconds (ms) and iterations per second (iter/s) are averaged across 100 runs.}
\label{tab:inf}
}
\end{table}
We implemented an 8-bit quantized integer-only LSTM with PWL model based on a custom PyTorch \cite{paszke2019pytorch} fork from 1.7.1. We implemented an integer-only PWL kernel using NEON intrinsics. We benchmark the models on a Huawei P20 Pro smartphone,  using the $\mathrm{speed\_benchmark\_torch}$ tool\footnote{\url{https://github.com/pytorch/pytorch/blob/1.7/binaries/speed_benchmark_torch.cc}}. We warm up each model for five runs, measure the inference time a hundred times, and report an average. The sequence length used is 128, and the batch size is one. We benchmark our iRNN LSTM model using PWLs with 32 pieces and eight pieces, which achieve up to $2\times$ speedup. We also evaluate our iRNN with full-precision computations (iRNN w/o QAct) for the activation where no speedup was observed for this state size. We believe it is due to round-trip conversions between floating points and integers (Table \ref{tab:inf}). There is much room for improvements to achieve even greater speedup, such as writing a C++ integer-only LSTM cell, fusing operations, and better PWL kernel implementation.
  
\begin{table*}
\begin{center}
\begin{tabular}{ l | c  c }
%\hline
	  Unimplemented Operations  & val & test\\
	  \hline
	  
      Quantized Weights (Full-precision) & $60.27 \pm 0.34$ & $58.02 \pm 0.34$ \\
	  PWL & $60.40 \pm 0.05$ & $ 57.90 \pm 0.01$  \\
    Quantized Activations & $60.40 \pm 0.03$ & $57.95 \pm 0.003$ \\ 
      Quantized Element-wise ops& $60.08 \pm 0.10$ & $57.61 \pm 0.23$ \\
      Quantized Matmul & $60.10 \pm 0.05$ & $57.64 \pm 0.10$ \\ 
\end{tabular} 
\end{center}
\caption{Ablation study on quantized Mogrifier LSTM training on WikiText2. iRNN without PWL is the quantized model using LUT instead of PWL to compute the activation function. Best results are averaged across three runs, and standard deviations are reported.}
\label{table:mogrifier-ablation}
\end{table*}

\begin{table*}
\begin{center}
\begin{tabular}{ c | c  c }
%      \hline 
	  Full-precision model & val & test\\
	  \hline
	  LayerNorm LSTM & $98.58 \pm 0.35$ & $ 94.84 \pm 0.21$ \\
      MadNorm LSTM & $\mathbf{97.20} \pm 0.47$ & $\mathbf{93.63} \pm 0.74$ \\ 
\end{tabular} 
%}
\end{center}
\caption{Word-level perplexities on PTB for a full-precision LSTM with LayerNorm and a full-precision model with MadNorm.  Best results are averaged across three runs, and standard deviations are reported.}
\label{table:ptb-fp-ln-madnorm}
\end{table*}

\section{Conclusion}

Speech and text applications are increasingly being deployed in IoT edge applications (i.e. ASR in autonomous vehicles or mobile devices, neural machine translation on portable translator devices, or even in many of today's smart home appliances). As RNN models form the basis of most these applications, it is vital that efficient, hardware-friendly models are designed to ensure usability on the edge. This requires that models use low-precision computation which can improve device latency, power usage and memory. 

In this text, we propose a task-agnostic and flexible methodology which enables integer-only RNNs that can tackle this problem. We are the first to offer an approach to quantize all existing operations in modern RNNs, supporting normalization and attention with a little performance drop. We evaluated our approach on high-performance LSTM-based models for language modeling and ASR, which have distinct architectures and variable computational requirements. Our experiments showed promising results on popular architectures and datasets namely; language modeling on WikiText2 with Mogrifier LSTM where our integer-only Mogrifier LSTM achieved similar performance to the full-precision model with only 0.3 perplexity increase with a PWL of 32 pieces, as well as ASR on LibriSpeech with ESPRESSO LSTM where our quantized solution only saw a $1.25$ WER\% drop compared to the full-precision model. These experiments sucessfully show that RNNs can be fully quantized while achieving similar accuracy as their full-precision counterpart. 

To emphasize the benefit of our integer-only RNN models on end-user devices, we benchmark our iRNN LSTM on a Huawei P20 Pro smartphone, where we obtain up to $2\times$ inference speedup and $4\times$ memory reduction. This implementation showed that even a simple integer-only model can see performance benefit compared to a full-precision while maintaining good accuracy. 

The method proposed in this text, highlights that it is now possible to deploy a wide range of accurate and high performing RNN-based applications on the edge. This opens the door for more improved usability and accessibility to advanced speech and text processing applications for the end-user.

\begin{appendices}

\section{}
\label{sec:appendix-A}
\subsection{Details on LSTM-based models}
\label{sec:appendix-lstm-continue}
For BiLSTM cells, nothing stated in section Integer-only LSTM network is changed except that we enforce the forward LSTM hidden state $\overrightarrow{\mathbf{h}}_t$ and the backward LSTM hidden state $\overleftarrow{\mathbf{h}}_t$ to share the same quantization parameters so that they can be concatenated as a vector. If the model has embedding layers, they are quantized to 8-bit as we found they were not sensitive to quantization. If the model has residual connections (e.g., between LSTM cells), they are quantized to 8-bit integers. In encoder-decoder models, the attention layers would be quantized using the method described in Section \ref{sec:qattention}. The last fully-connected layer weights of the model are 8-bit quantized to allow for 8-bit matrix multiplication. We do not quantize the outputs and let them remain 32-bit integers as often this is where it is considered that the model has done its job and that some postprocessing is performed (e.g., beam search).

\subsection{Experimental details}

%The model's last fully-connected layer's weights are 8-bit quantized to allow for 8-bit matrix multiplication. However, we do not quantize the outputs and let them remain 32-bit integers as often this is where it is considered that the model has done its job and that some postprocessing is performed (e.g. beam search). 

%\subsection{Experimental details}

\begin{table*}
\begin{center}
\begin{tabular}{ c | c  c  c  c  c }
    %   \hline 
%      \hline
	   & parameters & standard  & weight   & activation  & Total\\
	   &&training & quant.& quant. & GPU time\\
	  \hline
	  LM PTB & 4M & 1h & 13h & 10h & 24h  \\
      LM WikiText2 & 44M & 37h & 87h & 50h & 174h \\
      ASR LibriSpeech & 174M & 144h & 36h & 12h & 192h
\end{tabular} 
%}
\end{center}
\caption{Details about the model parameters and training time for each different experiment setup: language modeling on PTB (LM PTB), language modeling on WikiText2 (LM WikiText2), automatic speech recognition on LibriSpeech (ASR LibriSpeech). Parameters are given in the order of millions. We report the total number of V100 GPU hours as well as the GPU hours used for each different training phase, i) pre-trained model (standard training), ii) quantized model without PWL (weight quant.), iii) quantized model with PWL activation quantization (activation quant.) and the total GPU time.}
\label{table:model-info}
\end{table*}

In this section, we provide further details of our experimental setups.  The number of parameters and training time of the model are reported in Table \ref{table:model-info}.

% \begin{table*}
% \begin{center}
% \begin{tabular}{ l | c | c }
% \hline
% 	  iRNN Mogrifier LSTM & val & test\\
% 	  \hline
	  
% 	  w/o PWL & $60.40 \pm 0.05$ & $ 57.90 \pm 0.01$  \\
%     ~~w/o Quantized Activations & $60.40 \pm 0.03$ & $57.95 \pm 0.003$ \\ 
%       ~~~w/o Quantized Element-wise ops& $60.08 \pm 0.10$ & $57.61 \pm 0.23$ \\
%       ~~~~w/o Quantized Matmul & $60.10 \pm 0.05$ & $57.64 \pm 0.10$ \\ 
%       ~~~~~w/o Quantized Weights (Full-precision) & $60.27 \pm 0.34$ & $58.02 \pm 0.34$ \\
% \end{tabular} 
% \end{center}
% \caption{Ablation study on quantized Mogrifier LSTM training on WikiText2. iRNN w/o PWL is the quantized model using LUT instead of PWL to compute the activation function. Best results are averaged across 3 runs, and standard deviations are reported.}
% \label{table:mogrifier-ablation}
% \end{table*}

\subsubsection{LayerNorm LSTM on PTB}
\label{sec:appendix-ptb}
Preprocessing of the dataset was performed \cite{mikolov2012subword}. The vocabulary size is 10K. We report the best \emph{perplexity} per word on the validation set and test set for a language model of embedding size 200 with one LayerNormLSTM cell of state size 200. The lower the perplexity, the better the model performs. These experiments focus on the relative increase of perplexity between the full-precision models and their 8-bit quantized counterparts. We did not aim to reproduce state-of-the-art performance on PTB and went with a na\"ive set of hyper-parameters. The full-precision network is trained for 100 epochs with batch size 20 and BPTT \cite{werbos1990bptt} window size of 35. We used the SGD optimizer with weight decay of $10^{-5}$ and learning rate 20, which is divided by 4 when the loss plateaus for more than two epochs without a relative decrease of $10^{-4}$ in perplexity. We use gradient clipping of 0.25. We initialize the quantized models from the best full-precision checkpoint and train for another 100 epochs. We did not enable quantization to gather range statistics to compute the quantization parameters for the first five epochs.

\subsubsection{Mogrifier LSTM on WikiText2}
\label{sec:appendix-mogrifier}
We describe the experimental setup for Mogrifier LSTM on WikiText2. Note that we follow the setup of \cite{melis2020mogrifier} where they do not use dynamic evaluation \cite{krause2018dyneval} nor Monte Carlo dropout \cite{gal2016mcdropout}.
The vocabulary size is 33279. We use a two-layer Mogrifier LSTM with embedding dimension 272, state dimension 1366, and capped input gates. We use six modulation rounds per Mogrifier layer with low-rank dimension 48. We use 2 Mixture-of-Softmax layers \cite{yang2018mos}. The input and output embedding are tied. We use a batch size of 64 and a BPTT window size of 70. We train the full-precision Mogrifier LSTM for 340 epochs, after which we enable Stochastic Weight Averaging (SWA) \cite{izmailov2018swa} for 70 epochs. For the optimizer we used Adam \cite{kingma2014adam} with a learning rate of $\approx 3\times 10^{-3}$, $\beta_1=0$, $\beta_2=0.999$ and weight decay $\approx 1.8\times 10^{-4}$. We clip gradients' norm to 10. We use the same hyper-parameters for the quantized models from which we initialize with a pre-trained full-precision and continue to train for 200 epochs. During the first two epochs, we do not perform QAT, but we gather min and max statistics in the network to have a correct starting estimate of the quantization parameters. After that, we enable 8-bit QAT on every component of the Mogrifier LSTM: weights, matrix multiplications, element-wise operations, and activations. Then we replace activation functions in the model with quantization-aware PWLs and continue training for 100 epochs.  
We perform complete ablation on our method to study the effect of each component. Quantizing the weights and matrix multiplications cover about $0.1$ of the perplexity increase. There is a clear performance drop after adding quantization of element-wise operations with an increase in the perplexity of about $0.3$. This is due to element-wise operations in the cell and hidden state computations affecting the flow of information across timesteps and the residual connections across layers. Adding quantization of the activation does not impact the performance of the network.

\subsubsection{ESPRESSO LSTM on LibriSpeech}
\label{sec:appendix-asr}
 The encoder comprises 4 CNN-BatchNorm-ReLU blocks followed by 4 BiLSTM layers with 1024 units. The decoder consists of 3 LSTM layers of units 1024, with Bahdanau's attention on hidden states of the encoder and residual connections between each layer. The dataset preprocessing is precisely the same as in \cite{wang2019espresso}. 
 We train the model for 30 epochs on one V100 GPU, approximately six days to complete. We use a batch size of 24 while limiting the maximum number of tokens in a mini-batch to 26000. Adam is used with a starting learning rate of $0.001$, which is divided by 2 when the validation set metric plateaus without a relative decrease of $10^{-4}$ in performance. Cross-entropy with uniform label smoothing  $\alpha=0.1$ \cite{szegedy2016inceptionv3} is used as a loss function. The model predictions are weighted at evaluation time using a pre-trained full-precision 4-layer LSTM language model (shallow fusion). We consider this language model as an external component to the ESPRESSO LSTM; we do not quantize it due to the lack of resources. In our language modeling experiments, we have already shown that quantized language models retain their performance. We refer the reader to \cite{wang2019espresso} and training script\footnote{\url{https://github.com/freewym/espresso/blob/master/examples/asr_librispeech/run.sh}} for a complete description of the experimental setup.

 \begin{figure}
\centering
\begin{subfigure}[b]{0.45\linewidth}
\centering
\includegraphics[width=\linewidth]{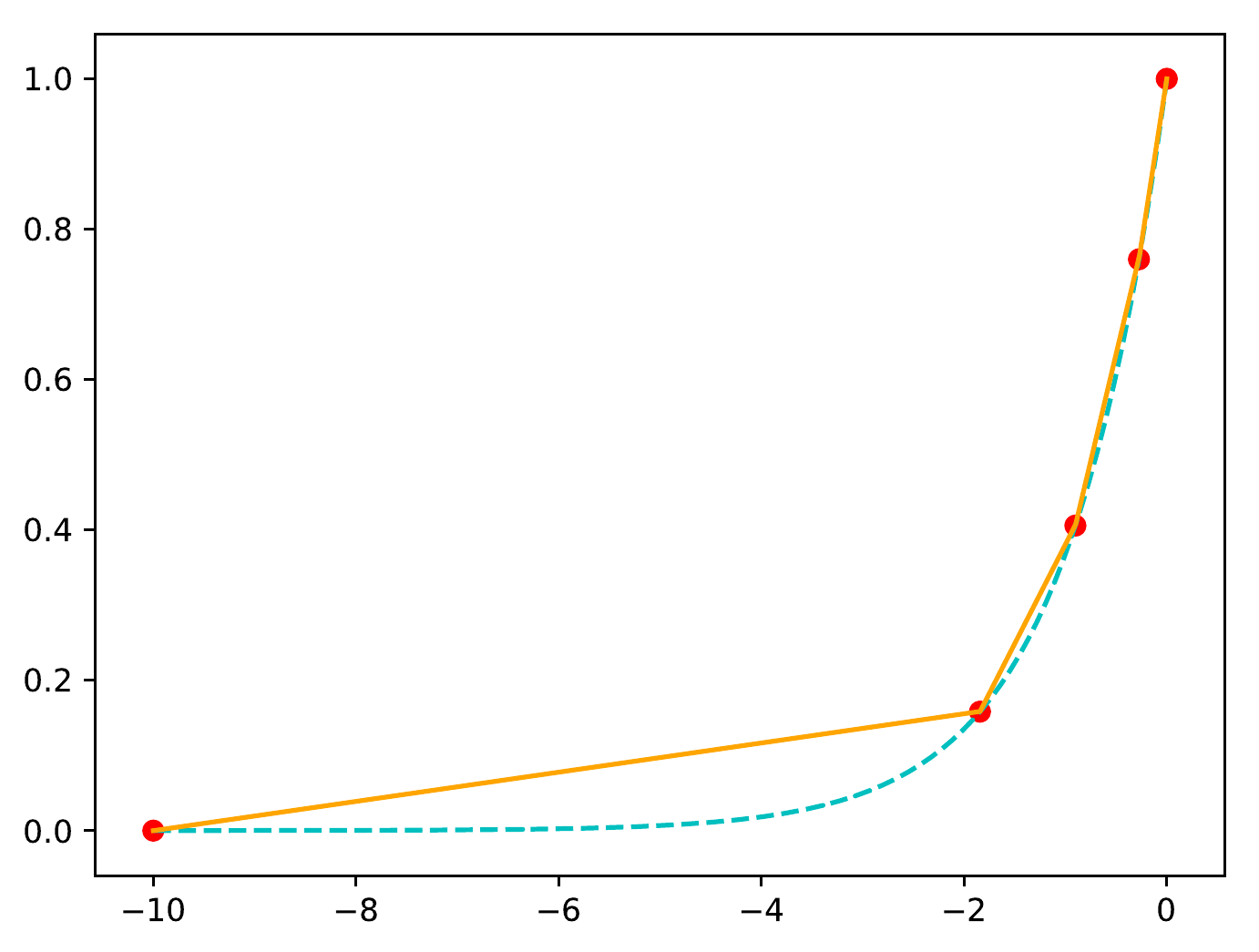}
\subcaption{}
\end{subfigure}
\begin{subfigure}[b]{0.45\linewidth}
\centering
\includegraphics[width=\linewidth]{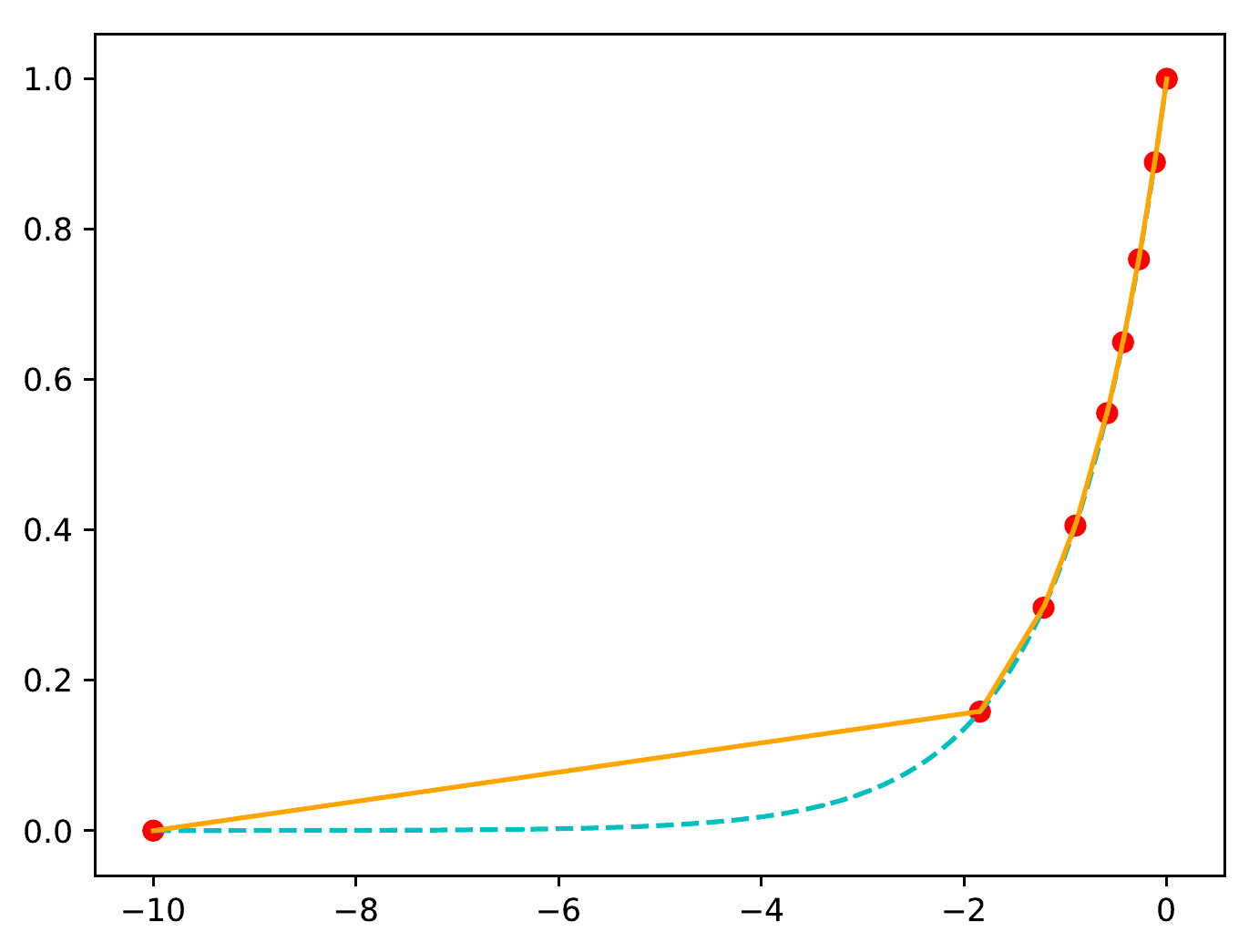}
\subcaption{}
\end{subfigure}\\
\begin{subfigure}[b]{0.45\linewidth}
\centering
\includegraphics[width=\linewidth]{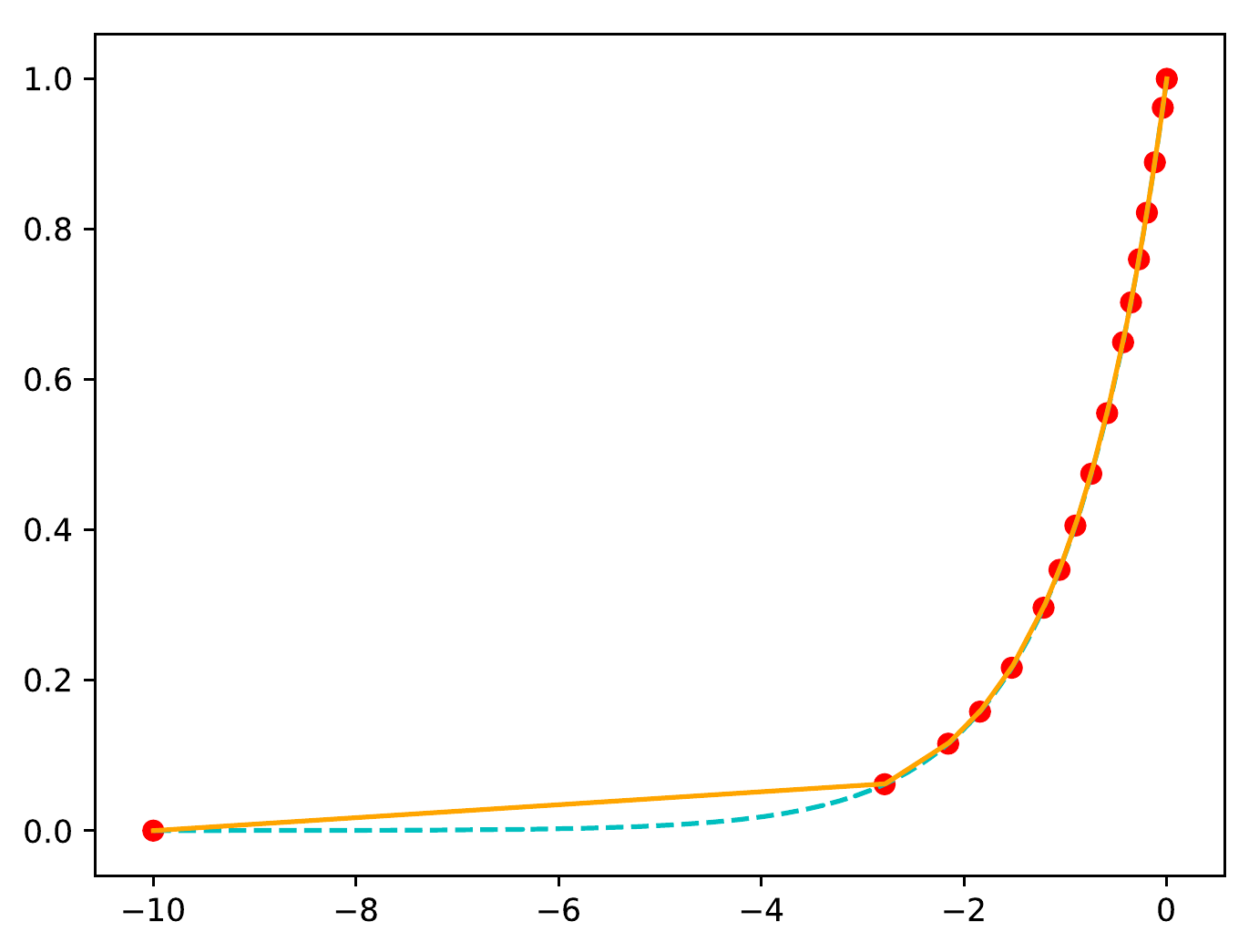}
\subcaption{}
\end{subfigure}
\begin{subfigure}[b]{0.45\linewidth}
\centering
\includegraphics[width=\linewidth]{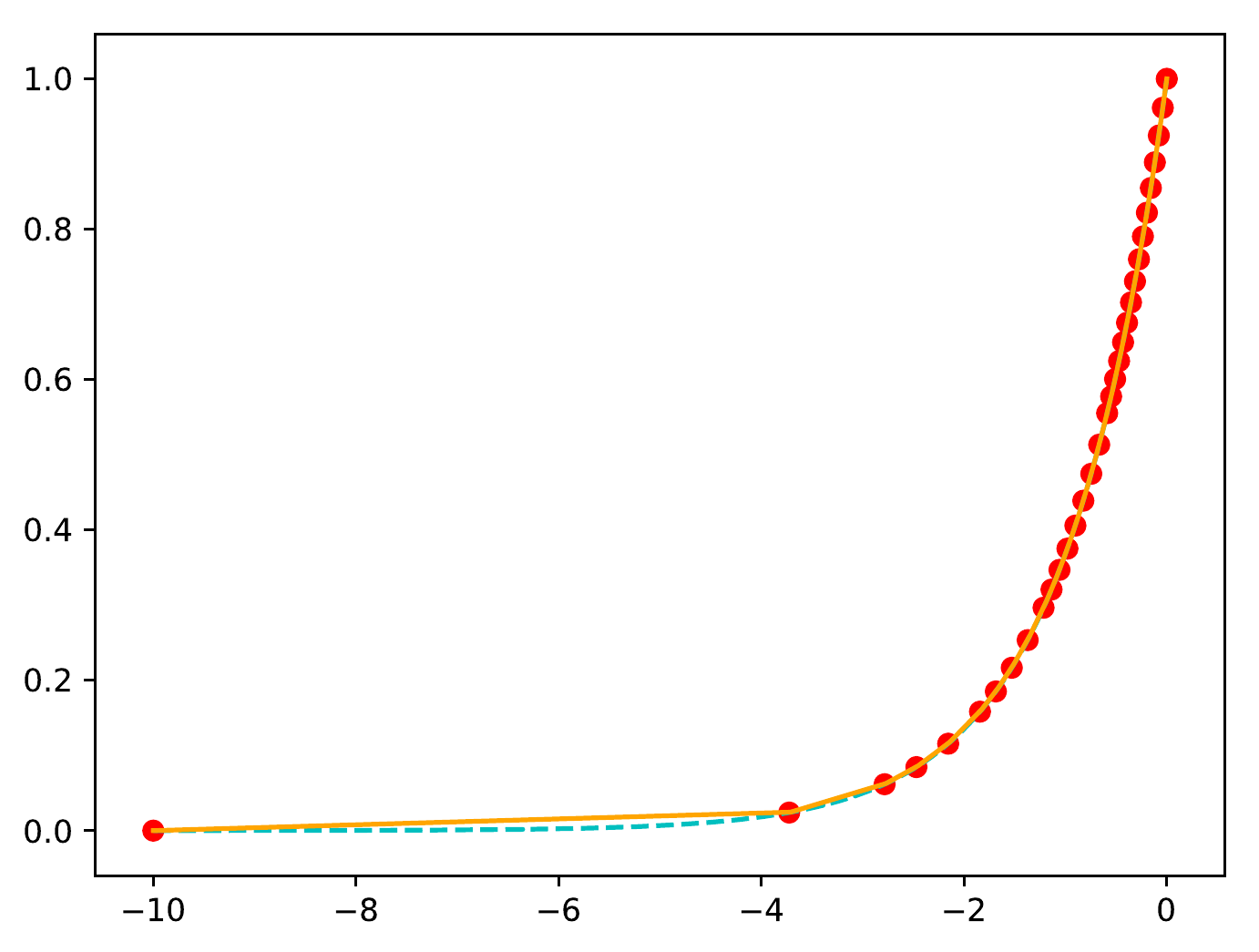}
\subcaption{}
\end{subfigure}\\
\centering
\begin{subfigure}[b]{0.45\linewidth}
\centering
\includegraphics[width=\linewidth]{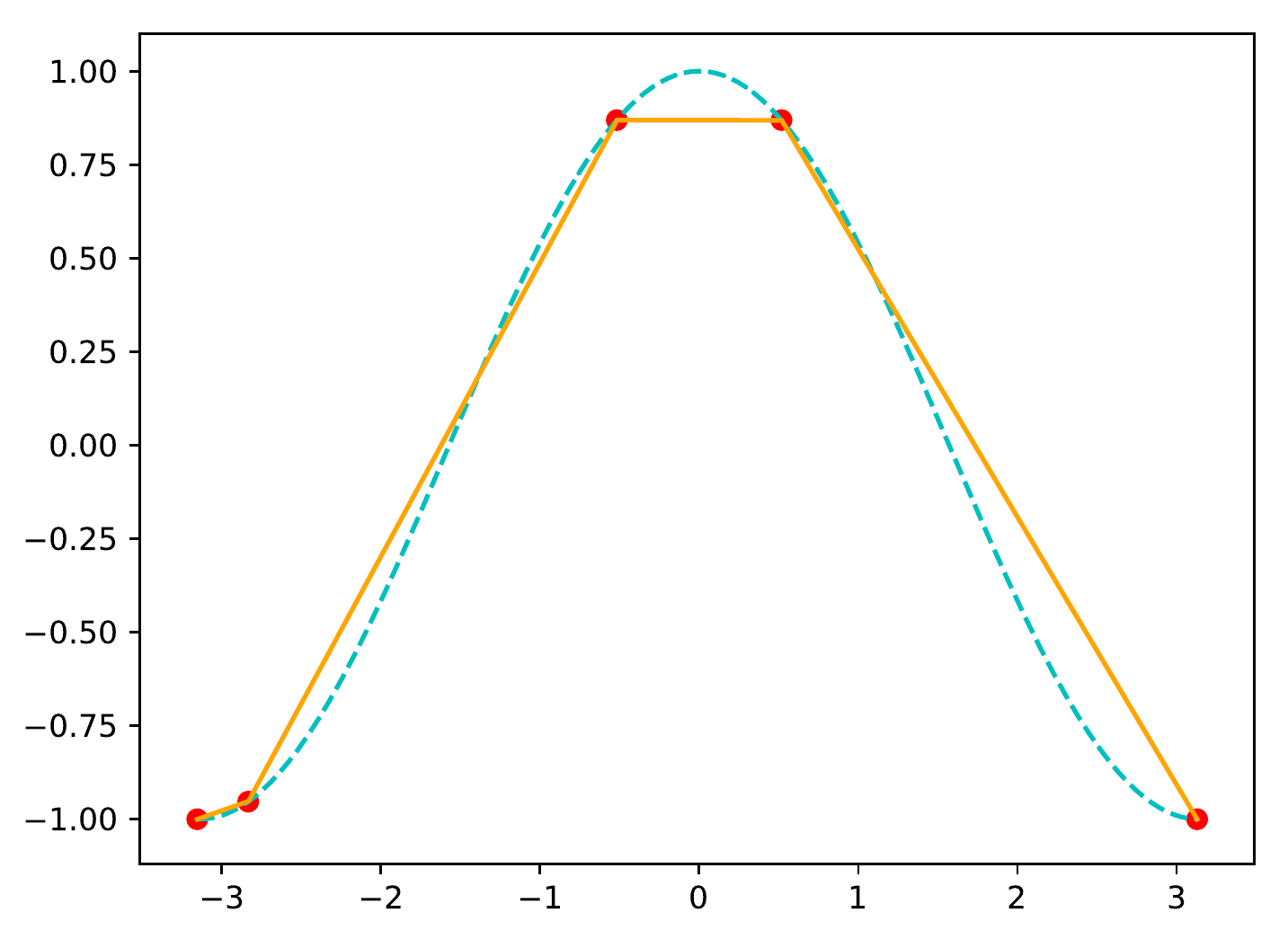}
\subcaption{}
\end{subfigure}
\begin{subfigure}[b]{0.45\linewidth}
\centering
\includegraphics[width=\linewidth]{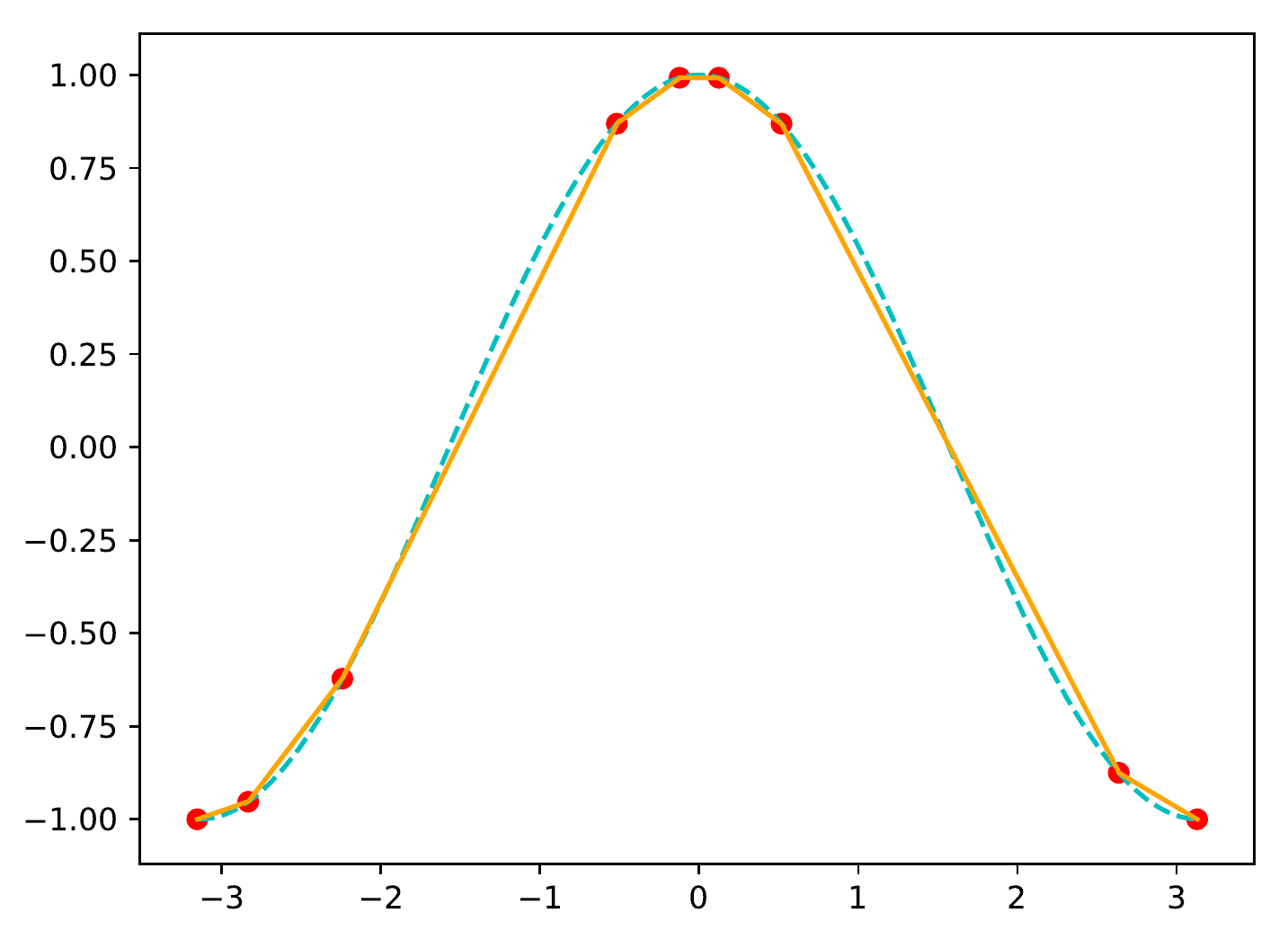}
\subcaption{}
\end{subfigure}\\
\begin{subfigure}[b]{0.45\linewidth}
\centering
\includegraphics[width=\linewidth]{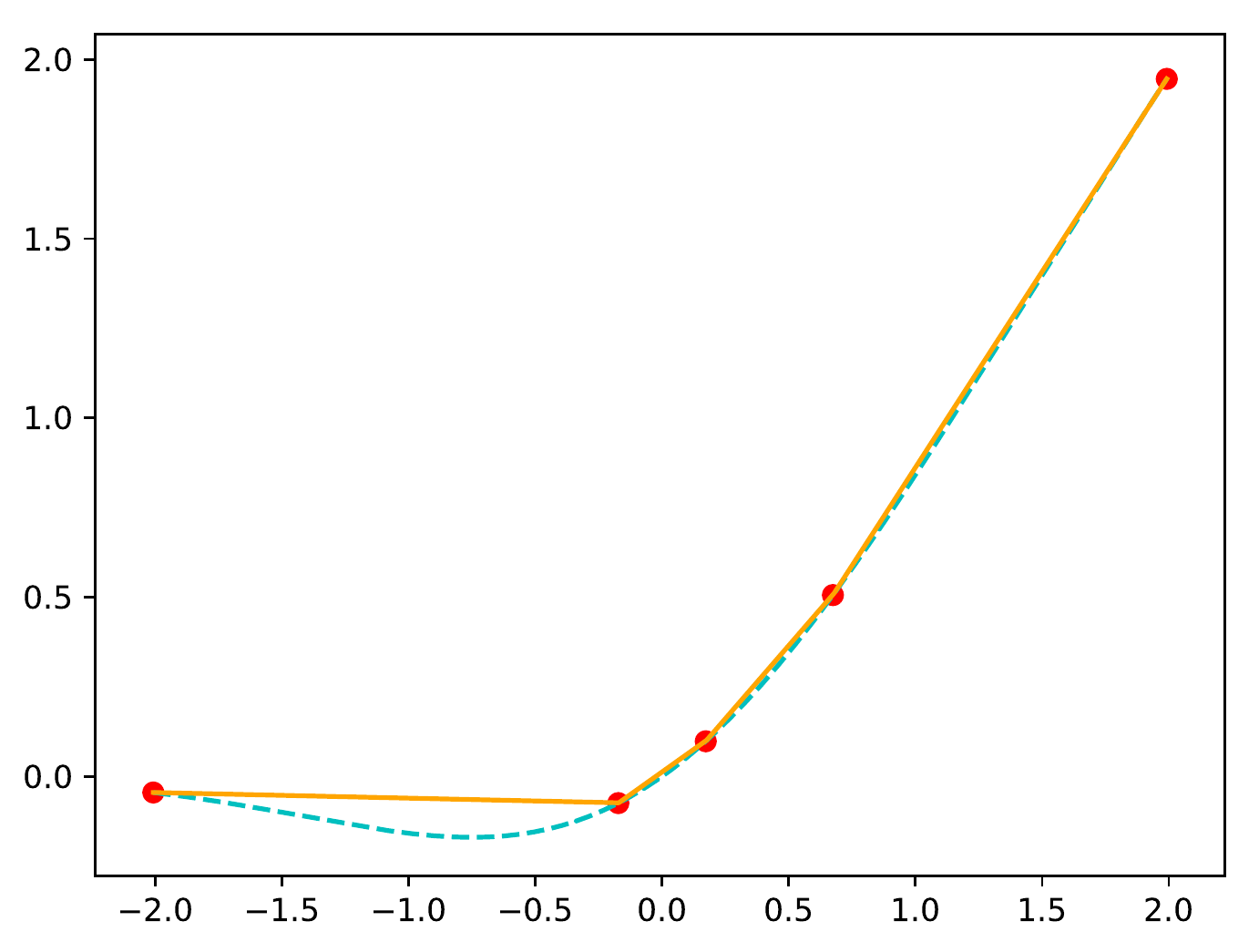}
\subcaption{}
\end{subfigure}
\begin{subfigure}[b]{0.45\linewidth}
\centering
\includegraphics[width=\linewidth]{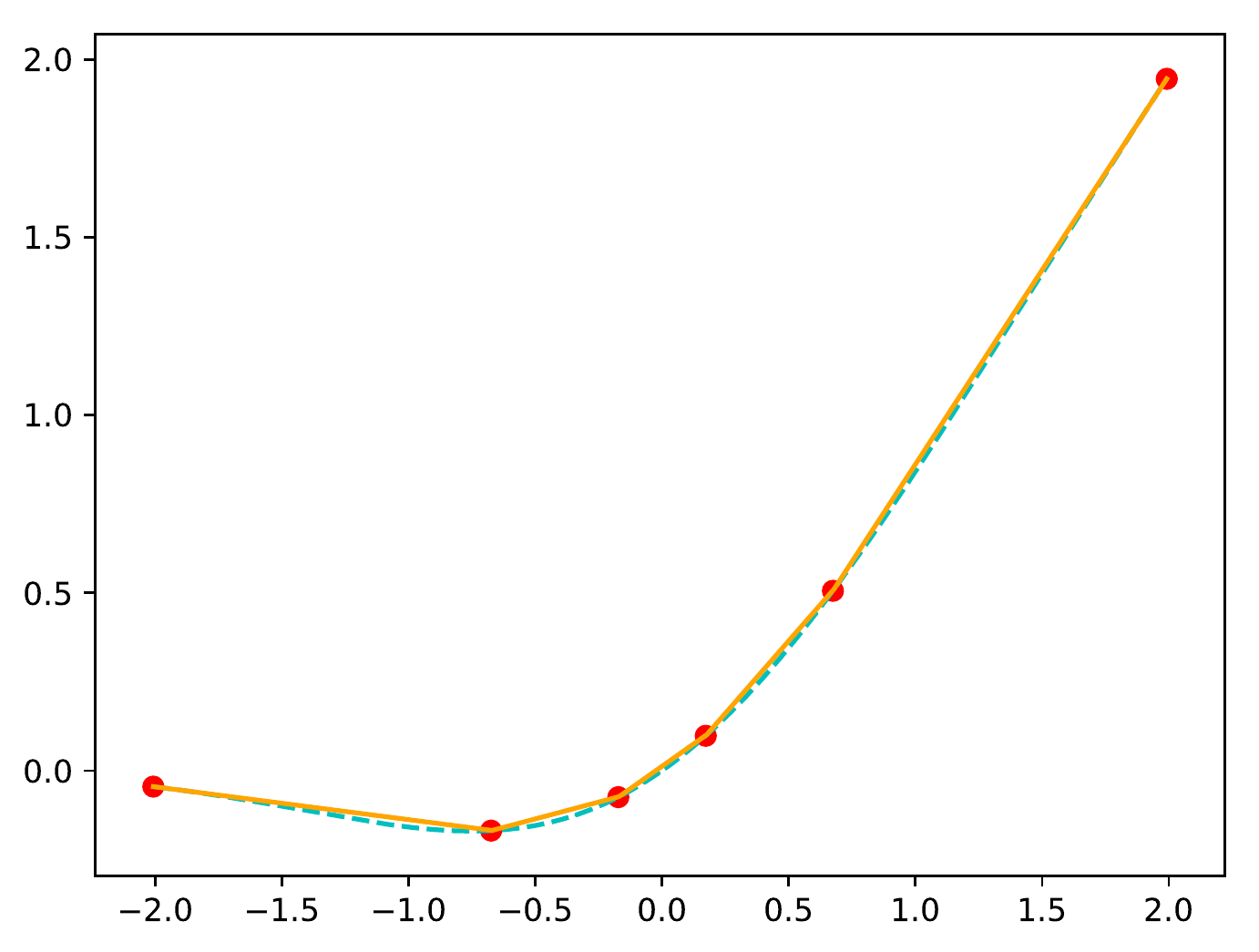}
\subcaption{}
\end{subfigure}
\caption{Quantization-aware PWL for several functions: exponential function approximation over $x \in [-10, 0]$ with a) 4, b) 8, c) 16, d) 32 pieces; cosine over $x \in [-\pi, \pi]$ with e) 4, f) 8 pieces; GeLU~\cite{hendrycks2016gelu} over $x \in [-2, 2]$ with g) 4, h) 5 pieces.}
\label{fig:other-pwl}
\end{figure}

We initialize the quantized model from the pre-trained full-precision ESPRESSO LSTM. Due to the lack of resources, we have trained the quantized model for only four epochs. The quantized model is trained on 6 V100 GPUs where each epoch takes two days, so a total of 48 GPU days. The batch size is set to 8 mini-batch per GPU with a maximum of 8600 tokens. We made these changes since, otherwise, the GPU would run out of VRAM due to the added fake quantization operations. For the first half of the first epoch, we gathered statistics for quantization parameters then we enabled QAT. The activation functions are swapped with quantization-aware PWL in the last epoch. The number of pieces for the quantization-aware PWLs is 96, except for the exponential function in the attention, which is 160, as we found out it was necessary to have more pieces because of its curvature. The number of pieces used is higher than that in the language modeling experiments. However, the difference is that the inputs to the activation functions are 16-bit rather than 8-bit, although the outputs are still quantized to 8-bit. It means we need more pieces to capture the input resolution better. Note that it would not be feasible to use a 16-bit Look-Up Table to compute the activation functions due to the size and cache misses, whereas using 96 pieces allows for a 170x reduction in memory consumption compared to LUT. 
%\begin{table}
% \begin{tabular}{ c | c | c }
%       \hline 
% 	  Full-precision model & val & test\\
% 	  \hline
% 	  LayerNorm LSTM & $98.58 \pm 0.35$ & $ 94.84 \pm 0.21$ \\
%       MadNorm LSTM & $\mathbf{97.20} \pm 0.47$ & $\mathbf{93.63} \pm 0.74$ \\ 
% \end{tabular} 
% \caption{Word-level perplexities on PTB for a full-precision LSTM with LayerNorm and a full-precision model with MadNorm.  Best results are averaged across 3 runs, and standard deviations are reported.}
% \label{table:ptb-fp-ln-madnorm}
% \end{table}

\begin{figure*}
\includegraphics[width=\textwidth]{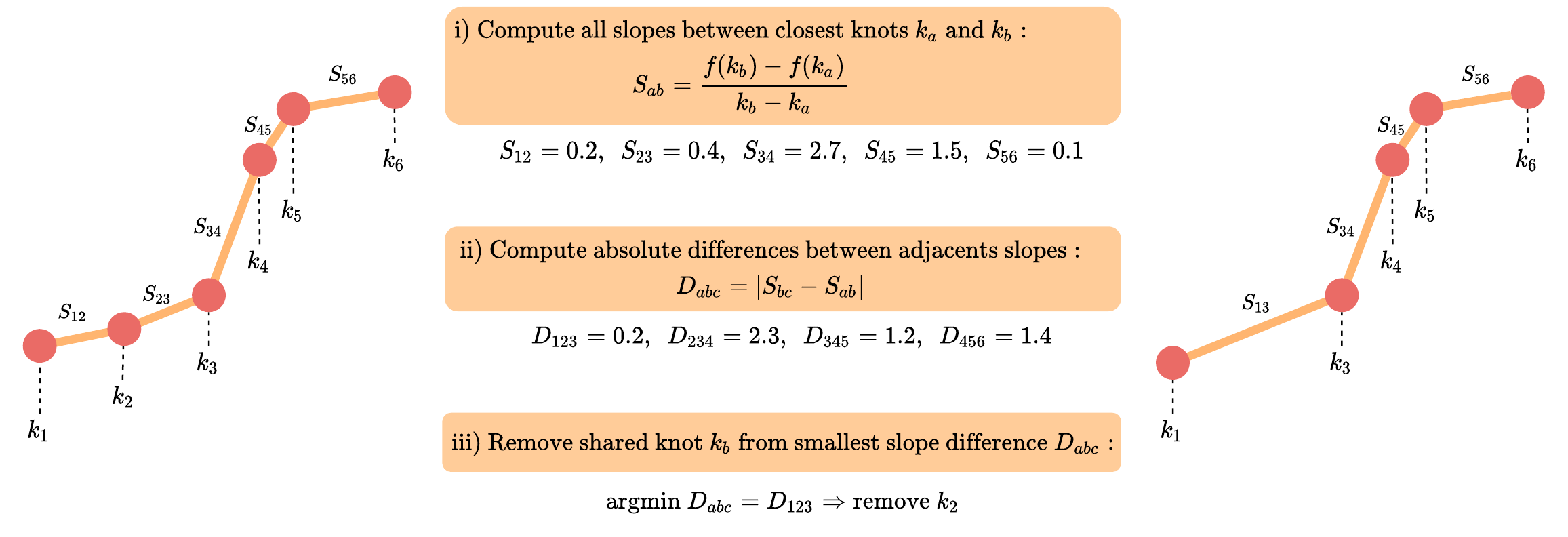}
\caption{Example of an iteration from our proposed quantization-aware PWL Algorithm \ref{alg:pwl}. The algorithm proceeds to reduce the number of pieces by merging two similar adjacents pieces. In this figure, the slopes $S_{12}$ and $S_{23}$ are the most similar pieces; therefore, the knot $k_2$ is removed.}
\label{fig:pwl-algo}
\end{figure*}

\section{}
\label{sec:appendix-quant}
\setcounter{table}{0}

The following section provides some examples of integer-only arithmetic and more details on fixed-point scaling. 
\subsection{Multiplication}
\label{appendix:qmul}
To illustrate how integer-only multiplication is achieved, we define an example utilizing (\ref{eq:quantmul}). Defining $u \in [u_{\min}, u_{\max}]=[-1, 1]$ and $w \in [w_{\min}, w_{\max}]=[0,5]$, the multiplication between two numbers from those ranges will fall into $[z_{\min}, z_{\max}]=[-5, 5]$. From (\ref{eq:scale-zero}), for 8-bit quantization, we have $S_u \approx 0.0078$, $Z_u=128$, $S_w \approx 0.0196, Z_w=0, S_z \approx 0.0392, Z_z=128$. Given $u=-0.8$ and $w=2.3$, we have $q_u=25$ and $q_w=117$. Therefore, following (\ref{eq:quantmul}),
\begin{align*}
    q_z &= \Big\lfloor \frac{S_u S_w}{S_z}\Big(q_u q_w - q_u Z_w - q_w Z_u + Z_u Z_w \Big) \Big\rceil + Z_z \label{eq:qmul-example}\nonumber\\ 
    & = \Big\lfloor \frac{0.0078 \times 0.0196}{0.0392}\Big(25 \times 117  - 25 \times 0 - 117 \times 128 + 128 \times 0 \Big) \Big\rceil + 128 \nonumber \\
    &= 81. %\label{eq:qmul-example-result}
\end{align*}
Using (\ref{eq:dequant}), the floating-point representation of $q_z$ is $r_z=-1.8424$ which is close to $uv=-1.8399$. Note that we lost precision at two levels, the first time when quantizing $u$ and $v$, then the second time when quantizing $z$, the multiplication output.

\subsection{Addition}
\label{sec:appendix-qadd}
As mentioned in Section \ref{sec:quant-operators}, addition with quantized numbers can take two forms. The first form is when the two numbers to be added share the same scaling factor and zero-point. For instance, given $x_1=-0.3, x_2=0.7$ from $[-1,1]$, and $S_x=0.0078, Z_x=128$, we have $q_{x_1}=90$ and $q_{x_2}=218$. The result value $y$ will fall into the range $[-2,2]$, therefore $S_y \approx 0.0157$ and $Z_y=128$. Then, because they share the same quantization parameters, following (\ref{eq:qadd-same}),

\begin{align*}
    q_y &= \Big\lfloor \frac{1}{S_y}\Big(S_x(q_{x_1} - Z_x) + S_x(q_{x_2} - Z_x)\Big) \Big\rceil + Z_y \\
    &= \Big\lfloor \frac{S_x}{S_y}\Big(q_{x_1} + q_{x_2} - 2Z_x\Big) \Big\rceil + Z_y \\
    &= \Big\lfloor \frac{0.0078}{0.0157}\Big(90 + 218 - 256\Big) \Big\rceil + 128 \\
    &= 154.
\end{align*} 

We have $r_y=0.4082$, while $x_1 + x_2 = 0.3999$. The second form is when the two numbers do not share the same scaling factor and zero-point. Define $a \in [a_{\min}, a_{\max}]=[-1, 1]$ and $b \in [b_{\min}, b_{\max}]=[0,5]$, the addition between two numbers from those ranges will fall into $[c_{\min}, c_{\max}]=[-1, 6]$. We get $S_a \approx 0.0078$, $Z_a=128$, $S_b \approx 0.0196, Z_b=0, S_c \approx 0.0274, Z_c=36$. For $a=-0.9$, $b=3.9$, we have $q_a=13$ and $q_b=199$. The quantized addition result $q_c$, following (\ref{eq:qadd-diff}), is,

\begin{align*}
    q_c &= \Big\lfloor \frac{S_a}{S_c}(q_a - Z_a) + \frac{S_b}{S_c}(q_b - Z_b)\Big\rceil + Z_c \\
    &= \Big\lfloor \frac{0.0078}{0.0274}(13 - 128) + \frac{0.0196}{0.0274}199\Big\rceil + 36 \\
    &= 146
\end{align*}

and $r_y=3.0140$ while $a+b=3.0$.

\subsection{Fixed point arithmetic}
\label{sec:appendix-fixed-point}
Even with the most careful rounding, fixed-point values represented with a scaling factor S may have an error of up to $\pm 0.5$ in the stored integer, that is, $\pm 0.5 S$ in the value. Therefore, smaller scaling factors generally produce more accurate results. On the other hand, a smaller scaling factor means a smaller range of values stored in a given program variable. The maximum fixed-point value that can be stored in a variable is the largest integer value that can be stored into it, multiplied by the scaling factor; and similarly for the minimum value. For example, Table \ref{tab:fp8}  gives the implied scaling factor $S$, the minimum and maximum representable values. The accuracy $\delta = S/2$ of values can be represented in 16-bit signed binary fixed-point format, depending on the number f of implied fraction bits.

 \begin{table}[]
     \centering
     \begin{tabular}{c c c c }
         scaling &  precision & signed (low, high) & unsigned (low, high) \\
         \hline
         $2^1$ & $2.0$ & $(-256,254)$ & $(0,510)$ \\
         $2^0$ & $1.0$ & $(-128,127)$ & $(0,255)$ \\
         $2^{-1}$ & $0.5$ & $(-64,63.5)$ & $(0,127.5)$ \\
         $2^{-2}$ & $0.25$ & $(-32,31.75)$ & $(0,63.75)$ \\
         $2^{-3}$ & $0.125$ & $(-16,15.875)$ & $(0,31.875)$ \\
         $2^{-4}$ & $0.0625$ & $(-8,7.9375)$ & $(0,15.9375)$ \\
         $2^{-5}$ & $0.03125$ & $(-4,3.96875)$ & $(0,7.96875)$ \\
         $2^{-6}$ & $0.015625$ & $(-2,1.984375)$ & $(0,3.984375)$ \\
         $2^{-7}$ & $0.0078125$ & $(-1,0.9921875)$ & $(0,1.9921875)$ \\
         $2^{-8}$ & $0.00390625$ & $(-0.5,0.49609375)$ & $(0,0.99609375)$ \\
     \end{tabular}
     \caption{Fixed point format with common signed scaling and bias format for 8 bit mantissa.}
     \label{tab:fp8}
 \end{table}

To convert a number from a floating-point to a fixed-point, one may divide it by the scaling factor S, then round the result to the nearest integer. Care must be taken to ensure that the result fits in the destination variable or register. Depending on the scaling factor and storage size, and the range of input numbers, the conversion may not entail any rounding. To convert a fixed-point number to floating-point, in contrast, one may convert the integer to floating-point and then multiply it by the scaling factor S. This conversion may entail rounding if the integer's absolute value is greater than 224 (for binary single-precision IEEE floating point) or of 253 (for double-precision). In addition, overflow or underflow may occur if $\abs S$ is very large or small. However, most computers with binary arithmetic have fast bit shift instructions that can multiply or divide an integer by any power of 2, particularly an arithmetic shift instruction. These instructions can be used to quickly change scaling factors that are powers of 2 while preserving the sign of the number.

\end{appendices}

\bibliography{main-rev01}% common bib file
%% if required, the content of .bbl file can be included here once bbl is generated
%%\input sn-article.bbl

%% Default %%
%%\input sn-sample-bib.tex%

\end{document}